%% file: main_new.tex
\pdfoutput=1
\documentclass{article}

\PassOptionsToPackage{numbers,compress}{natbib}
\usepackage[dblblindworkshop,final]{neurips_2026}
\workshoptitle{Foundation Models for Temporal Systems From Forecasting to World Modeling}

\usepackage[utf8]{inputenc}
\usepackage[T1]{fontenc}
\usepackage{hyperref}
\usepackage{url}
\usepackage{booktabs}
\usepackage{longtable}
\usepackage{array}
\usepackage{amsfonts}
\usepackage{amssymb}
\usepackage{nicefrac}
\usepackage{microtype}
\usepackage{xcolor}
\usepackage{graphicx}
\usepackage{xspace}
\usepackage{seqsplit}
\usepackage{wrapfig}
\usepackage{tikz}
\usepackage{todonotes}[inline]
\usepackage{adjustbox}
\usepackage[inline]{enumitem}
\usepackage{graphicx}
\newcommand{\pig}{\raisebox{-0.4ex}{\includegraphics[height=1.2em]{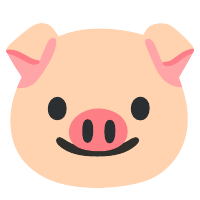}}}

\newcommand{\ds}[1]{\emph{\hspace{0pt}#1}}
\newcommand\comment[1]\null

\definecolor{plotblue}{HTML}{3464A8}
\definecolor{plotgrey}{HTML}{818181}

\graphicspath{{figures/}}
\hypersetup{colorlinks=true, linkcolor=black, citecolor=blue!55!black, urlcolor=blue!55!black}

\input{numbers}           %

\input{tables/symbols}    %

\newcommand{\browserurl}{\url{https://automl.github.io/ucr-provenance-audit/}}

\title{Fifty-Two Shades of Pig \pig: What the UCR Archive Really Asks Classifiers to Do}

\newcommand{\equal}{\textsuperscript{\rm *}}
\newcommand{\equaltext}{\equal\ Equal contribution}

\newcommand{\ufr}{\textsuperscript{\rm 1}}
\newcommand{\ufrc}{\textsuperscript{\rm 1,}}
\newcommand{\ufrtext}{\ufr\ University of Freiburg}

\newcommand{\prior}{\textsuperscript{\rm 2}}
\newcommand{\priorc}{\textsuperscript{\rm 2,}}
\newcommand{\priortext}{\prior\ Prior Labs}

\newcommand{\ellis}{\textsuperscript{\rm 3}}
\newcommand{\ellisc}{\textsuperscript{\rm 3,}}
\newcommand{\ellistext}{\ellis\ ELLIS Institute Tübingen}

\author{
Omar Swelam \ufrc\equal \quad Jaris Küken \ufrc\equal \quad Shi Bin Hoo \ufrc\equal\\ \textbf{Frank Hutter} \priorc\ellisc\ufr \quad \textbf{Lennart Purucker}\priorc\ufr \\ 
\ufrtext, \\ \priortext, \ellistext \\
\equaltext \\
\texttt{\{swelamo,kuekenj,hoos\}@cs.uni-freiburg.de}
}

\begin{document}

\maketitle

\begin{abstract}
The pigs were bleeding. That was the point of the experiment: could airway pressure reveal hemorrhage before it became critical? In the UCR Time Series Classification Archive, the de facto benchmark for Time Series Classification (TSC), the same recordings ask a different question: \emph{which of 52 pigs produced this trace?} The dataset \emph{PigAirwayPressure} is not an outlier. Although the archive is the standard benchmark, whether its tasks resemble the problems TSC is meant to solve has rarely been examined. We find that several datasets are repurposed from their original task, built from toy setups, or have flawed test splits. We further question how representative the archive's ordered non-temporal sequences (e.g. spectra, shape outlines) are of real TSC, and the potential downsides from the decomposition of multivariate problems. Our findings call for more careful interpretation of archive-wide scores and greater attention to deployment-relevant evaluation. The audit is at \url{https://github.com/automl/ucr-provenance-audit}.

\end{abstract}

\section{Introduction}
A benchmark does more than rank algorithms: it defines what a research community counts as progress. Meaningful progress requires the benchmarks tasks to resemble the problems encountered in practice~\citep{raji2021everything, koch2021reduced}. TSC is applied in diverse domains such as clinical monitoring~\citep{clifford2017af}, human activity recognition~\citep{anguita2013har}, and satellite remote sensing~\citep{pelletier2019tempcnn}, and a realistic benchmark should reflect this breadth. Diversity alone, however, does not make a benchmark representative: each task must also be posed as it would be in practice, with labels and evaluation protocols that correspond to a real-world problem.

The UCR archive has become the de facto standard benchmark for univariate TSC. %
Successive generations of distance-based, ensemble, convolutional, deep and representation-learning methods are compared on it \citep{dempster2020rocket,middlehurst2021hc2,yue2022ts2vec,middlehurst2024redux}, as are recent pretrained models~\citep{goswami2024moment,feofanov2025mantis,kueken2026timee}.Yet many of the archives datasets are knowingly not reflective of real world applications~\citep{dau2019ucr}. %
It admits several ordered non-temporal sequences, such as image outlines or spectra, and is biased towards image-derived datasets and datasets shaped by research interest~\citep{dau2019ucr}. While reasonable for a diverse and reproducible testbed, these choices are at odds with the archive's common use as a benchmark of realistic TSC.

We revisit the UCR archive from a provenance perspective, tracing each of its 128 datasets to its source to examine how each task was derived from the original data.
\footnote{The full audit, with evidence for every verdict, is browsable at \browserurl.}
We identify two recurring design choices that weaken the link between archive performance and real-world TSC: \textbf{artificial task construction} and \textbf{train/test leakage}. We further raise open questions about \textbf{pre-segmented and aligned inputs}, \textbf{ordered but non-temporal sequences}, and \textbf{multivariate recordings decomposed into several univariate datasets}.
We argue that the community-wide adoption of UCR in its current form as the benchmark for univariate TSC, together with the absence of more realistic alternatives, calls for action towards evaluation settings that reflect real-world TSC problems.

\section{Background and Related Work}
\textbf{Time Series Classification Archives.}
The UCR archive~\citep{dau2019ucr} and its multivariate counterpart, the UEA archive~\citep{bagnall2018uea}, are the standard benchmarks for TSC. Together they comprise 158 tasks, 128 univariate and 30 multivariate, drawn from image outlines, spectroscopy and simulated processes alongside ECG, EEG, motion capture and industrial sensor recordings. Recent collections follow the same pattern at larger scale: MONSTER~\citep{dempster2025monster} adds substantially larger datasets, and Multiverse~\citep{middlehurst2026multiverse} consolidates the UEA archive, MONSTER and several domain-specific collections into 133 multivariate problems. The bake-off series~\citep{bagnall2016bakeoff,middlehurst2024redux} imposed rigor on how comparisons over these archives are run, but likewise takes the task collection as given. Prior critiques target properties of the archive's tasks: \citet{hu2016streaming} document the simplifications of the pre-segmented setting, \citet{dhariyal2023backtobasics} show that tabular classifiers are competitive on many tasks, and \citet{zhang2025revisit} show that temporal order is often irrelevant and the methods can emphasize on a fixed specific position rather than patterns. Each identifies a symptom without asking why the archive contains such tasks in the first place.

\textbf{Benchmark Construction.}
Recent work on AI evaluation distinguishes a benchmark as an instrument for separating algorithms from a benchmark as a proxy for downstream applicability, and warns against reading aggregate scores on convenience-driven collections as general capability~\citep{raji2021everything}. Neighboring fields have responded with explicit curation: TabArena~\citep{erickson2025tabarena} admits only datasets that originate from real prediction problems and excludes derived or artificial tasks; RelBench~\citep{robinson2024relbench} defines tasks directly on relational databases with temporally valid splits. TSC has no equivalent. Existing critiques diagnose properties of particular datasets; but we are not aware of any analysis for the origin of the tasks these datasets were collected for nor whether they represent realistic TSC applications. Our provenance analysis is a first step towards such an examination.

\section{A Provenance Audit of UCR: Two Ways Its Tasks Fail as Benchmarks} \label{sec:audit}

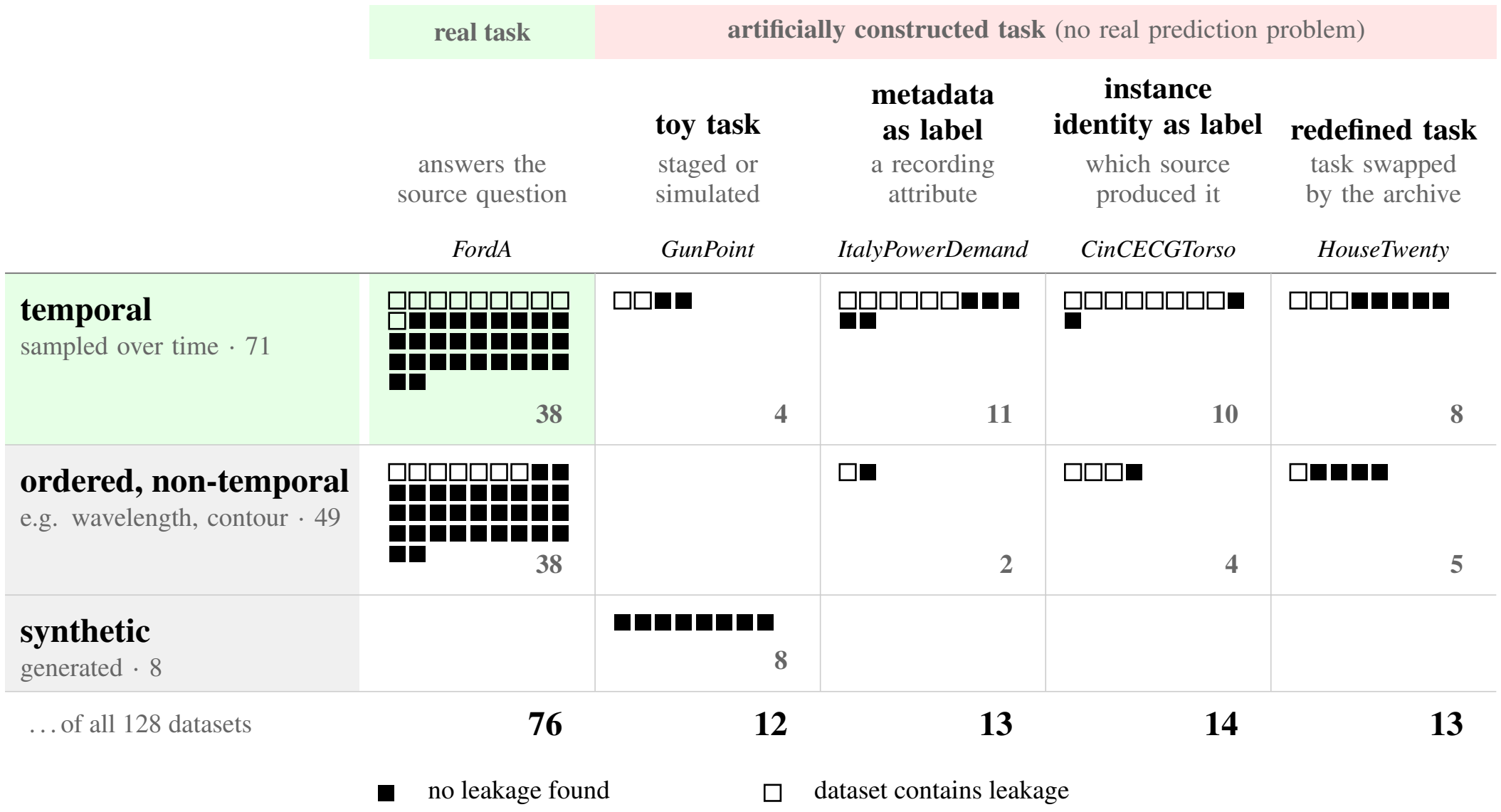
\begin{figure}[!t]
  \centering
    \begin{adjustbox}{width=1.0\columnwidth}
    \input{figures/composition-figure}
  \end{adjustbox}
  \vspace{-16pt}
  \caption{We show the archive datasets distribution over the different realism criterion. Our audit identifies only $38/128$ UCR datasets as real temporal tasks; about a quarter of these are prone to split leakage. We indicate the proportion of datasets failing our leakage criteria (white boxes) and with no leakage found (black boxes) within each category, although most splits cannot be checked (53 of the 76 real-task datasets are \emph{Unclear}).}
  \label{fig:composition}
  \vspace{-9pt}
\end{figure}

We assess all 128 UCR datasets through three stages: 1) the source study that produced the data, 2) the construction of the labelled task from the data, and 3) the construction of the train/test split. Appendix~\ref{app:protocol} details the protocol.

Two questions decide a dataset's standing as a benchmark task and organize this section. A dataset fails on task construction (Section~\ref{sec:construction}) if the source study posed no real task or the archive constructed the label itself. It fails on leakage (Section~\ref{sec:leakage}) if the test set shares series or subjects with the training set. We further flag 65 datasets where the subjects or session identity are underspecified, so the leakage counts below are lower bounds. Whether the series index is physical time is recorded but not judged, and we return to it in Section~\ref{sec:open-questions}. Figure~\ref{fig:composition} summarizes the outcome of our realism provenance.

\subsection{Artificial Task Construction}\label{sec:construction}
For 52 of the 128 UCR datasets, the classification target is not a real-world prediction problem.
Some labels differ from the source question or were constructed post hoc where no classification task was posed. Other source tasks were built only to exercise algorithms, with no application behind the label.

While such constructed labels are best understood as pretext tasks in the sense of self-supervised learning, where pretext accuracy is a known-unreliable proxy for downstream performance~\citep{Kolesnikov_2019_CVPR}, UCR reports the pretext task as the benchmark result itself.
We identify four mechanisms by which such labels are constructed:

\textbf{Metadata as label} (13): the archive derives classes from a recording attribute.
\emph{ItalyPowerDemand} labels demand curves by season (October--March vs.\ April--September); \emph{ECGFiveDays} labels one subject's ECG segments by which of two recording dates they come from.

\textbf{Instance identity as label} (14): the archive uses the recording's source as the class, regardless of what the source study asked.
In \emph{CinCECGTorso}, the 2007 challenge concerned infarct location and extent~\citep{physionet2007challenge}, but the archive labels each ECG by which of four subjects produced it;
the three \emph{Pig*} datasets turn a bleeding-detection study~\citep{TSCPigs} into a 52-way subject-identification problem.

\textbf{Task replacement} (13): the archive replaces the source's label with a different task, or defines classes by its own rule where the source posed none.
\emph{AllGestureWiimoteX/Y/Z} were collected for user identification but are benchmarked as gesture recognition \citep{allgesturewiimote};
\emph{HouseTwenty} takes a smart-meter recording of one home~\citep{murray2017refit} and asks whether a window is the whole-house signal or the sum of two appliances, a question the study never posed.

\textbf{Toy source tasks} (17): the data were simulated or staged by method developers, and the classes were chosen to demonstrate a method rather than to answer any external question. \emph{ShapeletSim} embeds one of two shapes in noise so that shapelet methods are guaranteed to separate them~\citep{bagnall2016bakeoff}; GunPoint has two actors draw a gun or point at UCR, staged in 2003 by the archive's curators~\citep{dau2019ucr}.

These mechanisms are not exclusive. For example, in \emph{GunPointMaleVersusFemale}, the recording was staged by the curators, and the archive then labels it by which of the two actors is performing rather than by the gun/point distinction the staging was designed around. 
Appendix~\ref{app:labels} lists them all with their original label (if one exists) and the constructed label, and marks the overlaps.

For a benchmark, the consequence is that accuracy on these tasks is a pretext score with no downstream target to validate it against: a method can rank first archive-wide by exploiting recording artifacts, subject identity or generator mechanics, none of which are the capabilities a practitioner deploying a TSC model needs.

\subsection{Train/Test Leakage}\label{sec:leakage}

For 41 of the 128 UCR datasets, the test set is not independent of the training set. 
This happens in two ways: identical series appear in both train and test splits, or the split itself introduces leakage by grouping (or not grouping) the subjects. Additionally in 65 datasets, the information needed to check source separation is missing. To our knowledge, no systematic leakage check was run by the curators or the donors. We flag two main concerning setups:

\begin{wrapfigure}{r}{0.45\linewidth}
  \centering
  \includegraphics[width=\linewidth,trim=28 32 25 27,clip]{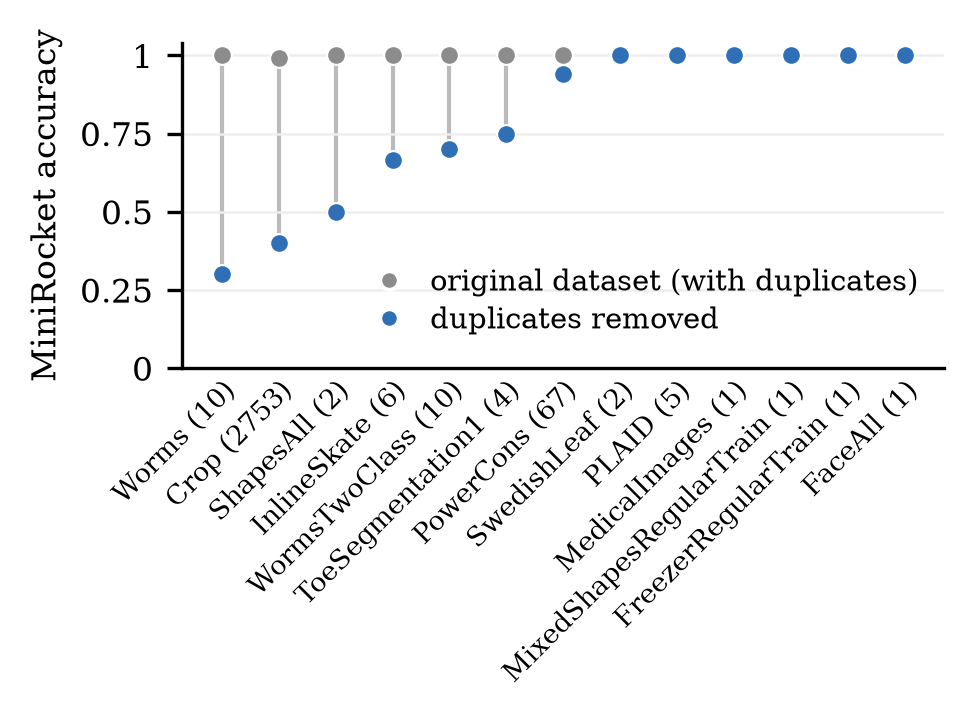}
  \caption{MiniRocket accuracy with the original training set (with duplicates, \textcolor{plotgrey}{grey}) and after removing the duplicates (\textcolor{plotblue}{blue}). On several datasets, duplication inflates test accuracy.}
  \label{fig:leakage}
  \vspace{-10pt}
\end{wrapfigure}

\textbf{Duplication} (17 datasets): at least one test series is value-identical to a training series; in \emph{PowerCons} this affects 37\% of the test set ($67/180$), in \emph{Crop} 16\% ($2753/16800$), in \emph{Worms} 13\% ($10/77$). Same-label exact matches permit a classifier to reproduce a training label without demonstrating generalisation to a new case (see Figure~\ref{fig:leakage} and Table~\ref{tab:duplicates}).

\textbf{Shared Sources} (24 datasets):  whether a split is valid depends on the task. 
Class recognition, whose task is to predict a class for subjects unseen in training, requires the split to be grouped by subject; instance recognition, whose task is to re-identify a known subject, requires the test set to contain new recordings of that subject.
On these 24 datasets one of the two rules is violated: in \ds{SemgHandMovementCh2} (class recognition) the grasps of all five subjects appear on both sides of the split;
in \ds{SemgHandSubjectCh2} (instance recognition) each of the five subjects was recorded in a single session, and the repetitions of that session are split by a stratified shuffle. %

These pose trustworthiness challenges, as scores should state which evaluation setting they represent in benchmarking. Further details can be found in Appendix~\ref{app:split}.

\section{Open Questions} \label{sec:open-questions}
\textbf{Do ordered sequences translate to temporal axes?}
The UCR archive encompasses multiple ordered datasets that span domains like spectrography and image contours. While one can argue that they provide meaningful tests for sequence models, temporal notions as warping, trends, periodicity and regime changes have no direct counterparts on a spatial or spectral axis. Treating both kinds of data as one benchmark therefore assumes a transfer the archive never tests\citep{bagnall2016bakeoff,middlehurst2024redux}.

\citet{zhang2025revisit} provide indirect evidence: applying a single fixed permutation to every series, destroying local order while leaving each value at a fixed position, barely changes accuracy on roughly a third of the 112 equal-length UCR datasets, so those tasks are solved from aligned values rather than from order, a strategy unavailable in deployment for TSC.

\textbf{Should decomposed multivariate recordings count as separate datasets?}
Extracting one channel from a multivariate recording is a useful test of a univariate model. Counting several channels as independent datasets is a different decision: they share subjects, labels, acquisition and the underlying phenomenon \citep{dau2019ucr,tscarchive_descriptions}. Archive-wide comparisons, including rank-based tests, treat datasets as independent samples of TSC problems; derived channels violate that assumption, and one source problem receives several correlated votes in the aggregate score.

\textbf{How much of the problem does preprocessing solve?}
Every UCR series is a pre-extracted instance, and in many datasets instances are further aligned to a common reference before release, such as image outlines to a canonical starting point. The archive therefore never asks where in a stream a class-defining pattern occurs or how long it lasts. \citet{hu2016streaming} argued that this pre-segmented setting is far removed from how classification is deployed on streams, yet the archive and the methods evaluated on it have continued to treat it as given.

The consequence is that, once patterns sit at near-fixed positions, time steps become features. This offers an explanation for two recent findings: tabular classifiers are competitive on many UCR tasks \citep{dhariyal2023backtobasics}, and disrupting temporal order often does not meaningfully affect predictive performance \citep{zhang2025revisit}. Whether performance on such tasks says anything about a method's ability to locate patterns in time, a key capability required in deployment, remains an open question.

\section{Conclusion}
The UCR archive made univariate TSC evaluation accessible and reproducible, and they remain a reasonable instrument for comparing algorithms. Our audit, however, shows that archive-wide performance should not be interpreted uncritically as performance on real-world TSC problems. Constructed targets, train/test dependencies, preprocessing choices, non-temporal sequences and repeated source problems each break that link, and they are common enough in UCR that an aggregate ranking cannot be read as a measure of deployment readiness. We argue for evaluation that complements the archives with provenance-aware valid task selection, deployment-relevant splits and capturing the realism of deployment environments to close the gap of explored caveats. Progress requires both better scores and benchmarks that measure the intended capability.

\section*{Acknowledgements}
  Funded by the European Union. Views and opinions expressed are however those of the author(s) only and do not necessarily reflect those of the European Union or the European Commission. Neither the European Union nor the European Commission can be held responsible for them. This work was supported by the European Union's Horizon Europe research and innovation program under grant agreement No 101214398 (ELLIOT). L.P. acknowledges funding
by the Deutsche Forschungsgemeinschaft (DFG, German Research Foundation) under SFB 1597
(SmallData), grant number 499552394; F.H. acknowledges the financial support of the Hector Foundation.
  \begin{center}
    \begin{minipage}{0.2\textwidth}
      \centering
      \includegraphics[width=\textwidth]{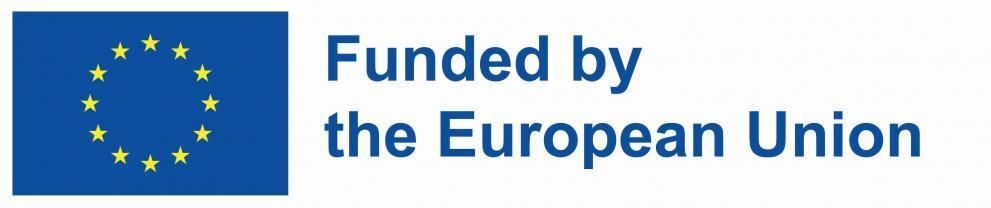}
    \end{minipage}
    \begin{minipage}{0.2\textwidth}
      \centering
      \includegraphics[width=\textwidth]{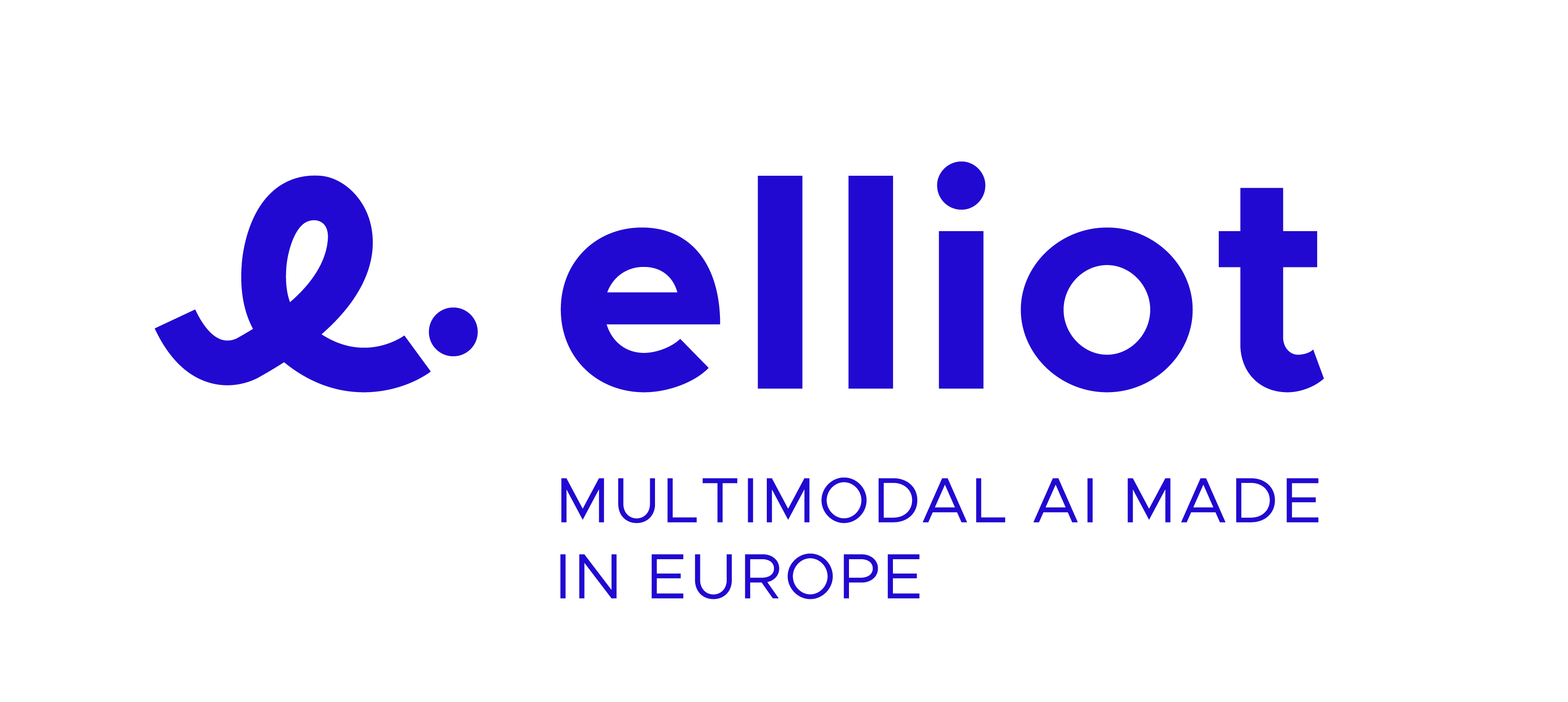}
    \end{minipage}
  \end{center}

\bibliographystyle{plainnat}
{\small\bibliography{references_revised}}

\clearpage
\appendix
\input{appendix}

\end{document}

%% file: numbers.tex
\providecommand{\xspace}{}

\newcommand{\accDupElectricDevices}{0.365\xspace}  %
\newcommand{\accRestElectricDevices}{0.752\xspace}  %
\newcommand{\accDupStrawberry}{0.000\xspace}  %
\newcommand{\accRestStrawberry}{0.986\xspace}  %

%% file: tables/symbols.tex
\newcolumntype{P}[1]{>{\raggedright\arraybackslash}p{#1}}

%% file: figures/composition-figure.tex
\begin{tikzpicture}[
  x=1cm, y=1cm,
  sq/.style={draw=black, line width=0.5pt},
  hd/.style={font=\footnotesize\bfseries, anchor=south, align=center, text width=1.95cm, inner sep=1pt},
  gl/.style={font=\scriptsize, text=black!60, anchor=north, align=center, text width=1.95cm, inner sep=1pt},
  ex/.style={font=\fontsize{6.5}{7.5}\selectfont\itshape, anchor=base, align=center},
  rl/.style={font=\small, anchor=north west, inner sep=1pt},
  rg/.style={font=\scriptsize, text=black!60, anchor=north west, text width=3.15cm, inner sep=1pt},
  ct/.style={font=\scriptsize\bfseries, text=black!60, anchor=south east},
  tt/.style={font=\small\bfseries, anchor=east},
  lg/.style={font=\scriptsize, anchor=west},
  band/.style={font=\scriptsize, text=black!60},
]
\def\xz{3.4} \def\cw{2.1} \def\pad{0.18} \def\pitch{0.19} \def\sqs{0.15}
\def\rtA{0}    \def\rhA{1.6}
\def\rtB{-1.6} \def\rhB{1.4}
\def\rtC{-3.0} \def\rhC{0.9}
\def\gb{-3.9}
\colorlet{pass}{green!10}
\colorlet{fail}{red!10}
\colorlet{lens}{black!6}

\newcommand{\cell}[4]{%
  \ifnum#3>0
    \pgfmathtruncatemacro{\nmax}{#3-1}
    \foreach \k in {0,...,\nmax}{
      \pgfmathtruncatemacro{\r}{floor(\k/9)}
      \pgfmathtruncatemacro{\c}{mod(\k,9)}
      \pgfmathsetmacro{\px}{\xz+#1*\cw+\pad+\c*\pitch}
      \pgfmathsetmacro{\py}{#2-\pad-\r*\pitch}
      \ifnum\k<#4 \draw[sq] (\px,\py) rectangle ++(\sqs,-\sqs);
      \else \fill[black] (\px,\py) rectangle ++(\sqs,-\sqs); \fi
    }
  \fi
}
\newcommand{\cellcount}[3]{\node[ct] at (\xz+#1*\cw+\cw-\pad, #2+0.08) {#3};}

\fill[pass] (\xz, 2.00)    rectangle (\xz+\cw,   2.50);
\fill[fail] (\xz+\cw, 2.00) rectangle (\xz+5*\cw, 2.50);
\node[band] at (\xz+0.5*\cw, 2.25) {\textbf{real task}};
\node[band, align=center, text width=8.2cm] at (\xz+3*\cw, 2.25)
  {\textbf{artificially constructed task} (no real prediction problem)};
\foreach \i/\cname/\cdef/\exm in {
  0/{}/{answers the source question}/{FordA},
  1/{toy task}/{staged or simulated}/GunPoint,
  2/{metadata as label}/{a recording attribute}/ItalyPowerDemand,
  3/{instance identity as label}/{which source produced it}/CinCECGTorso,
  4/{redefined task}/{task swapped by the archive}/HouseTwenty}{
  \node[hd] at (\xz+\i*\cw+0.5*\cw, 1.2) {\cname};
  \node[gl] at (\xz+\i*\cw+0.5*\cw, 1.15) {\cdef};
  \node[ex] at (\xz+\i*\cw+0.5*\cw, 0.15) {\exm};
}

\fill[pass] (0,\rtA) rectangle (\xz-0.1, \rtA-\rhA);
\fill[lens] (0,\rtB) rectangle (\xz-0.1, \rtB-\rhB);
\fill[lens] (0,\rtC) rectangle (\xz-0.1, \gb);
\fill[pass] (\xz,\rtA) rectangle (\xz+\cw, \rtA-\rhA);

\draw[black!50, line width=0.4pt] (0,\rtA) -- (\xz+5*\cw,\rtA);
\draw[black!20, line width=0.3pt] (0,\rtB) -- (\xz+5*\cw,\rtB);
\draw[black!20, line width=0.3pt] (0,\rtC) -- (\xz+5*\cw,\rtC);
\draw[black!20, line width=0.3pt] (0,\gb)  -- (\xz+5*\cw,\gb);
\foreach \i in {1,...,4}{\draw[black!20, line width=0.3pt] (\xz+\i*\cw,\rtA) -- (\xz+\i*\cw,\gb);}

\node[rl] at (0.1,\rtA-0.18) {\textbf{temporal}};
\node[rg] at (0.1,\rtA-0.55) {sampled over time $\cdot$ 71};
\node[rl] at (0.1,\rtB-0.18) {\textbf{ordered, non-temporal}};
\node[rg] at (0.1,\rtB-0.55) {e.g. wavelength, contour $\cdot$ 49};
\node[rl] at (0.1,\rtC-0.18) {\textbf{synthetic}};
\node[rg] at (0.1,\rtC-0.55) {generated $\cdot$ 8};

\cell{0}{\rtA}{38}{10} \cellcount{0}{\rtB}{38}
\cell{1}{\rtA}{4}{2}   \cellcount{1}{\rtB}{4}
\cell{2}{\rtA}{11}{6}  \cellcount{2}{\rtB}{11}
\cell{3}{\rtA}{10}{8}  \cellcount{3}{\rtB}{10}
\cell{4}{\rtA}{8}{3}   \cellcount{4}{\rtB}{8}
\cell{0}{\rtB}{38}{7}  \cellcount{0}{\rtC}{38}
\cell{2}{\rtB}{2}{1}   \cellcount{2}{\rtC}{2}
\cell{3}{\rtB}{4}{3}   \cellcount{3}{\rtC}{4}
\cell{4}{\rtB}{5}{1}   \cellcount{4}{\rtC}{5}
\cell{1}{\rtC}{8}{0}   \cellcount{1}{\gb}{8}

\node[font=\scriptsize, text=black!60, anchor=west] at (0.1, \gb-0.3) {\dots of all 128 datasets};
\foreach \i/\n in {0/76, 1/12, 2/13, 3/14, 4/13}{\node[tt] at (\xz+\i*\cw+\cw-\pad, \gb-0.3) {\n};}

\fill[black] (\xz+0.08, \gb-1.02) rectangle ++(\sqs,\sqs);
\node[lg] at (\xz+0.42, \gb-0.94) {no leakage found};
\draw[sq]    (\xz+3.68, \gb-1.02) rectangle ++(\sqs,\sqs);
\node[lg] at (\xz+4.02, \gb-0.94) {dataset contains leakage};
\end{tikzpicture}

%% file: appendix.tex
\section{Provenance Audit Protocol}
\label{app:protocol}

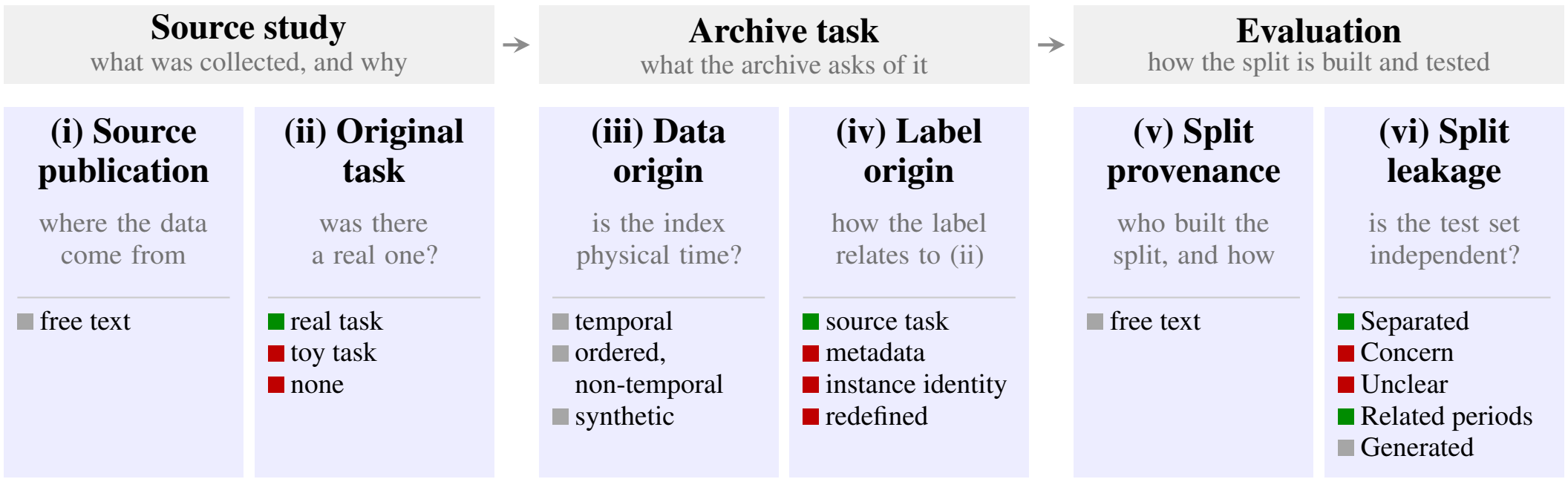
\begin{figure}[ht]
  \centering
  \input{figures/audit_pipeline}
  \caption{The six audit fields, two per stage of a dataset's life. Green marks the value that passes, red a value that fails or flags leakage, grey a value the verdict does not read.}
  \label{fig:pipeline}
\end{figure}

We describe how the six audit fields of Figure~\ref{fig:pipeline} were filled
for the 128 UCR datasets, which sources were admitted as evidence, how the checks on the data were run, and how agentic systems were used in the setup.

\subsection{The six fields}
\label{app:fields}

Each dataset is followed through three stages of its life, with two fields per stage
(Figure~\ref{fig:pipeline}). Fields (i) and (v) are free text and carry the evidence; the
other four take a value from a fixed vocabulary, and only those four feed the counts in
Section~\ref{sec:audit}.

\begin{description}\setlength{\itemsep}{1pt}
  \item[(i) Source publication.] The paper, thesis, competition or data release the
  series come from, with the passages read.
  \item[(ii) Original task.] Whether the source study posed a classification task of its
  own: \emph{real task} (a stakeholder outside the paper needed the prediction),
  \emph{toy task} (data staged or generated to demonstrate a method), or \emph{none}
  (the source collected data but posed no classification target).
  \item[(iii) Data origin.] What the series index is: \emph{temporal} (sampled over
  physical time), \emph{ordered, non-temporal} (wavelength, contour arc-length, spatial
  position, entity index), or \emph{synthetic} (a generator's step).
  \item[(iv) Label origin.] How the archive's class relates to (ii): \emph{source task}
  (the same target, possibly coarsened), \emph{metadata} (a recording attribute such as
  season, date, lead or device), \emph{instance identity} (which subject, sensor or
  actor produced the series), or \emph{redefined} (a target the archive defined where the
  source posed none or a different one).
  \item[(v) Split provenance.] Who built the default train/test split and by what rule,
  where this is documented at all.
  \item[(vi) Split leakage.] Whether the test set is independent of the training set:
  \emph{Separated} (no duplicates, and the unit a deployed model would face differs between the sets, as documented for the split or by construction because each series is a distinct event or object with no known grouping above that unit), \emph{Concern} (at least one test series is value-identical to a
  training series, or the unit a deployed model would face appears on both
  sides), \emph{Unclear} (the information needed to check separation of that unit is missing), \emph{Related periods} (different days or windows of the same
  ongoing process), or \emph{Generated} (synthetic tasks, not assessed by source separation).
\end{description}

A dataset fails on task construction (Section~\ref{sec:construction}) when (ii) is
\emph{toy task} or \emph{none}, or (iv) is \emph{metadata}, \emph{instance identity} or \emph{redefined}. It fails
on leakage (Section~\ref{sec:leakage}) when (vi) is \emph{Concern}.
\emph{Unclear} is never counted as a pass on leakage. The four datasets whose original task or label origin is \emph{undocumented} are counted with the real tasks. Field (iii) is recorded for every
dataset but does not enter either verdict (Section~\ref{sec:open-questions}).

\subsection{Admissible evidence}
\label{app:evidence}

Four kinds of evidence were admitted, and every value in the released audit carries a tag
saying which kind supports it.

\begin{itemize}\setlength{\itemsep}{1pt}
  \item \textbf{[S] Source text.} A passage of the source publication, the archive's own
  briefing document~\cite{dau2019ucr}, the donor's documentation, or the dataset page on
  \texttt{timeseriesclassification.com}, cited by section and page and quoted where the
  claim is decisive. Dead donor links were recovered through the Internet Archive where
  possible; a link that points to an unrelated resource is itself recorded as evidence of
  weak provenance.
  \item \textbf{[D] Shipped data.} A property computed on the archive's own files: the
  duplicate check of Appendix~\ref{app:leakage}, and one-line sanity checks used during
  drafting (sign of the values, whether the series length equals a
  count of entities in the source, whether split sizes reconcile with the source's
  arithmetic).
  \item \textbf{[R] Inference.} A conclusion the auditor drew from [S] or [D] rather than
  read in a source, such as identifying which of several challenges a vaguely described
  dataset comes from.
  \item \textbf{[I] Intended deployment.} An inference about intended deployment, such as
  which unit a deployed model would face.
\end{itemize}

Where the sources do not answer a question, the field is set to \emph{unknown} or
\emph{undocumented} and the record says what is missing. A fluent but unsupported account
of a dataset's origin was treated as a defect, not a filled field.

\subsection{Language-model assistance}
\label{app:llm}

The audit followed a five-stage pipeline: scripted scaffolding, agentic drafting, agentic review, an agentic fact-check of the released CSV against the records, and human verification. The prompts and protocols for each stage are released together with the audit.

\paragraph{Scaffolding.} A script scrapes the archive page of each dataset and fills the identity fields (split sizes, series length, number of classes, archive type tag). All judgment fields are left as placeholders.

\paragraph{Drafting.} An agent follows the archive's source links, reads the source publication, fills fields (i) to (v) according to the evidence rules of Appendix~\ref{app:evidence}, and runs the data checks. The protocol forbids filling fields from prior knowledge, requires the load-bearing sentence to be quoted with its location, and instructs the agent to lower the confidence field and state what is missing rather than infer unstated values.

\paragraph{Review.} A second, independent agentic pass, run under a separate protocol, fetches and reads every listed reference before inspecting the draft. For each decisive claim it records one of \emph{confirmed with quote}, \emph{quote mismatch}, \emph{not found}, \emph{contradicted}, or \emph{omitted by draft}, re-derives the field values from the verified evidence, and re-runs the duplicate check.

\paragraph{Fact-check.} Two independent agents (Claude Opus and GPT Astra) re-derived every verdict and fact column of the released CSV from the records, keeping the existing value where a record is silent, and a third agent reconciled the two.

\paragraph{Human verification.} The authors decided every verdict on which the agents disagreed or which the reconciling agent rated as low-confidence. For contested datasets, the authors re-read the cited sources themselves, and corrected records were re-checked.

\subsection{Leakage checks}
\label{app:leakage}

\paragraph{Duplicates.} For every dataset we loaded the shipped train and test files with
their original lengths and missing values, and hashed each series as a byte string of its
64-bit values with missing values mapped to a fixed sentinel, so that identical padding
hashes identically. Two series are duplicates when their hashes agree. We report, per
dataset, the number of test series whose values appear in the training set, the number of
distinct series overall, the duplicate-group size histogram, and whether any duplicate group
carries more than one label. 
\paragraph{Group overlap.} Duplicates are not the only way a test set can depend on the training set.
When the same subject, device or recording session contributes series to both sides of the split, a model can pass by recognizing the source rather than the class. We marked a split as leaking when the archive, the source or the released data establish such sharing, as clean when they document that it is absent, or when each series is a distinct event or object with no known grouping above it, and as unknown otherwise.

\subsection{Release}
\label{app:release}

The repository \url{https://github.com/automl/ucr-provenance-audit} contains the audit CSV with a column guide, one YAML record per dataset, the drafting, review and fact-check protocols, the duplicate-check script with its output, and the browser page at \browserurl. %

\clearpage
\section{Artificial Task Construction}\label{app:labels}
Tables~\ref{tab:labels-metadata}–\ref{tab:labels-toy} list all 52 datasets that fail task construction, because their label differs from the source task or the source task is a toy, grouped by the mechanism described in Section~\ref{sec:construction}. There are 57 entries for 52 datasets; five datasets fall under two mechanisms and appear in both tables. The column ``Source target or purpose'' states the question the source study stposed, or notes that no classification target was documented. A derived label does not by itself make a task invalid; the tables record what is being evaluated and who defined it.
{\footnotesize\setlength{\tabcolsep}{3pt}
\subsection{Metadata as Label (13)}\input{tables/labels_metadata}
\subsection{Instance Identity as Label (14)}\input{tables/labels_identity}
\clearpage\subsection{Task Replacement, Derivation or Subsetting (13)}\input{tables/labels_replacement}
\clearpage\subsection{Toy Source Tasks (17)}\input{tables/labels_toy}}

\clearpage
\section{Split Provenance}\label{app:split} 
The 41 UCR datasets assigned the \texttt{Concern} label, comprise 17 with at least one exact cross-split match and 24 more with a documented or data-verified shared source but no exact match. 

\subsection{Train/Test leakage or inconsistent labels}\label{app:duplicates}      

Table~\ref{tab:duplicates} reports the datasets with the duplicates and split matches. A test
series whose exact values already appear in training can be answered by recall rather than
by generalisation, so the two accuracy columns split each test set in two: \emph{acc.\
dup.} covers the test series that have an identical training counterpart, \emph{acc.\
rest} the remainder. The gap between them is large and one-directional. MiniRocket answers the repeated cases almost perfectly while scoring far lower on the rest of the same test set. Two datasets run the other way, \emph{ElectricDevices} (\accDupElectricDevices against \accRestElectricDevices) and \emph{Strawberry} (\accDupStrawberry against \accRestStrawberry), because some repeated series carry conflicting labels and no classifier can satisfy every copy.
  {\footnotesize\setlength{\tabcolsep}{3pt}\input{tables/duplicates}}

  \subsection{Split Concerns Against Evaluation}
  Table~\ref{tab:splits} examines whether each split supports the intended evaluation. A new instance means a case from a new source at the relevant level: for example, a new patient, a new physical sample or a fresh recording of a familiar person. Another window from the same recording is a different case, but does not provide a new recording occasion. We assign one verdict per dataset:       
  \begin{itemize}[leftmargin=*,itemsep=2pt,topsep=3pt]                                                   
  \item \emph{Concern} (41 datasets) identifies either a relevant source unit shared across the split that the archive, its description paper, the source study or the released data establish, or at least one test series value-identical to a training series. Documented source sharing can let accuracy benefit from familiar recording or sample characteristics, while same-label exact matches permit memorisation.
  \item \emph{Unclear} (65 datasets) means that the information needed to check separation is missing. 

  \item \emph{Separated} (7 datasets) means that the relevant sources differ between the sets, as documented or by construction when each series is a distinct event or object with no known grouping above that unit, supporting evaluation at that level. Separate sessions of familiar users, for example, do not establish performance on new users.

  \item \emph{Related periods} (7 datasets) identifies different days or windows from the same ongoing process, which may share seasons, events or operating conditions. Random assignment does not establish performance in a later period.

  \item \emph{Generated} (8 datasets) marks tasks not assessed here by patient, sample or recording separation. Their independence must instead be checked from how the examples were generated.

  \end{itemize}

  Overall, 106 of 128 datasets (82.8\%) either receive a \texttt{Concern} label because of a documented or data-verified shared source or an exact cross-split match, or lack sufficient information to verify source separation. Each verdict applies to the setting described in its row and qualifies what the score can be taken to demonstrate. Rows marked ``same as'' share the named explanation.

  {\footnotesize\setlength{\tabcolsep}{3pt}\input{tables/split_audit}}

%% file: figures/audit_pipeline.tex
\providecommand{\okm}{{\color{green!55!black}\rule{4pt}{4pt}}~}   %
\providecommand{\badm}{{\color{red!75!black}\rule{4pt}{4pt}}~}    %
\providecommand{\nam}{{\color{black!35}\rule{4pt}{4pt}}~}         %
\begin{tikzpicture}[x=1cm,y=1cm,
  >=stealth,
  every node/.style={inner sep=0pt,outer sep=0pt},
  stage/.style  ={fill=black!6},
  field/.style  ={fill=blue!7},
  name/.style   ={font=\small\bfseries, align=center},
  gloss/.style  ={font=\scriptsize, align=center, text=black!55},
  vals/.style   ={font=\scriptsize, align=left},
  arrow/.style  ={->,thick,black!45},
]
\def\W{13.9}    %
\def\SW{4.367}  \def\SG{0.4}    %
\def\FW{2.133}  \def\FG{0.1}    %

\foreach \s/\title/\sgloss in {
  0/Source study/{what was collected, and why},
  1/Archive task/{what the archive asks of it},
  2/Evaluation/{how the split is built and tested}} {
  \pgfmathsetmacro{\x}{\s*(\SW+\SG)}
  \fill[stage] (\x,6.45) rectangle ++(\SW,0.7);
  \node[name]  at (\x+\SW/2,6.93) {\title};
  \node[gloss] at (\x+\SW/2,6.63) {\sgloss};
}
\draw[arrow] (\SW+0.08,6.8) -- ++(\SG-0.16,0);
\draw[arrow] (2*\SW+\SG+0.08,6.8) -- ++(\SG-0.16,0);

\foreach \s/\f/\num/\fname/\fgloss/\fvals in {
  0/0/(i)/{Source\\publication}/{where the data come from}/{\nam free text},
  0/1/(ii)/{Original\\task}/{was there a real one?}/{\okm real task\\ \badm toy task\\ \badm none},
  1/0/(iii)/{Data\\origin}/{is the index physical time?}/{\nam temporal\\ \nam ordered,\\ \phantom{\nam}non-temporal\\ \nam synthetic},
  1/1/(iv)/{Label\\origin}/{how the label relates to (ii)}/{\okm source task\\ \badm metadata\\ \badm instance identity\\ \badm redefined},
  2/0/(v)/{Split\\provenance}/{who built the split, and how}/{\nam free text},
  2/1/(vi)/{Split\\leakage}/{is the test set independent?}/{\okm Separated\\ \badm Concern\\ \badm Unclear\\ \okm Related periods\\ \nam Generated}} {
  \pgfmathsetmacro{\x}{\s*(\SW+\SG)+\f*(\FW+\FG)}
  \fill[field] (\x,2.95) rectangle ++(\FW,3.3);
  \node[name,  anchor=north] at (\x+\FW/2,6.15) {\num~\fname};
  \node[gloss, anchor=north, text width=\FW cm-6pt] at (\x+\FW/2,5.3) {\fgloss};
  \draw[black!20] (\x+0.12,4.55) -- (\x+\FW-0.12,4.55);
  \node[vals,  anchor=north west] at (\x+0.12,4.43) {\fvals};
}

\end{tikzpicture}

%% file: tables/labels_metadata.tex
\begin{longtable}{@{}P{3.2cm} P{3.2cm} P{4.4cm} P{1.6cm}@{}}
\caption{Task-construction cases in the current audit: metadata. None means no classification target is documented, not that the source data had no scientific purpose. $^\dagger$ also listed under another mechanism.}\label{tab:labels-metadata} \\
\toprule
Dataset & Benchmark target & Source target or purpose & Label by \\
\midrule
\endfirsthead
\multicolumn{4}{l}{\emph{(continued)}} \\
\toprule
Dataset & Benchmark target & Source target or purpose & Label by \\
\midrule
\endhead
\midrule
\multicolumn{4}{r}{\emph{continues}} \\
\endfoot
\bottomrule
\endlastfoot
\ds{ItalyPowerDemand} & Cold season (Oct--Mar) vs warm season (Apr--Sep) & None; utility demand records & Curators \\
\ds{PowerCons} & Warm vs cold season & None; EDF consumption records & Curators \\
\ds{Chinatown} & Weekend vs weekday & None; Melbourne pedestrian counts & Curators \\
\ds{DodgerLoopDay} & Day of week & Unsupervised event detection; no class target & Curators \\
\ds{DodgerLoopGame} & Whether a Dodgers home game took place that day & Unsupervised detection of game traffic events; schedule used as validation ground truth & Curators \\
\ds{DodgerLoopWeekend} & Weekend vs weekday & As DodgerLoopDay & Curators \\
\ds{ECGFiveDays} & Which of two dates, five days apart & None documented; one subject's clinical ECG & Curators \\
\ds{GunPointOldVersusYoung}$^\dagger$ & Recording session: 2003 vs 2018 & Gun-draw vs pointing action & Collectors (=~curators) \\
\ds{SemgHandGenderCh2} & Sex of five subjects & Six hand-grasp types & Curators \\
\ds{TwoLeadECG} & Which of two ECG leads & None used; one long-term ECG record & Curators \\
\ds{Rock} & Four rock types & ASTER spectral-library metadata; no defined task & Curators \\
\ds{Worms} & Worm strain: wild type or one of four mutants & Unsupervised discovery of behavioural motifs; no class target & Curators \\
\ds{WormsTwoClass} & Wild type vs mutant & As Worms & Curators \\
\end{longtable}

%% file: tables/labels_identity.tex
\begin{longtable}{@{}P{3.2cm} P{3.2cm} P{4.4cm} P{1.6cm}@{}}
\caption{Task-construction cases in the current audit: identity. None means no classification target is documented, not that the source data had no scientific purpose. $^\dagger$ also listed under another mechanism.}\label{tab:labels-identity} \\
\toprule
Dataset & Benchmark target & Source target or purpose & Label by \\
\midrule
\endfirsthead
\multicolumn{4}{l}{\emph{(continued)}} \\
\toprule
Dataset & Benchmark target & Source target or purpose & Label by \\
\midrule
\endhead
\midrule
\multicolumn{4}{r}{\emph{continues}} \\
\endfoot
\bottomrule
\endlastfoot
\ds{PigAirwayPressure} & Which of 52 pigs & Before vs during induced bleeding & Curators \\
\ds{PigArtPressure} & Which of 52 pigs & As PigAirwayPressure & Curators \\
\ds{PigCVP} & Which of 52 pigs & As PigAirwayPressure & Curators \\
\ds{CinCECGTorso} & Which of four people & Infarct extent and location & Curators \\
\ds{MelbournePedestrian} & Which of ten sensor locations & Pedestrian monitoring; no class target & Curators \\
\ds{SemgHandSubjectCh2} & Which of five subjects & Six hand-grasp types & Curators \\
\ds{Yoga}$^\dagger$ & Which of two actors, from silhouettes & Actor discrimination, posed by the collectors as a demo task & Collectors (=~curators) \\
\ds{GunPointMaleVersusFemale}$^\dagger$ & Actor sex; confounded with identity (two actors) & Gun-draw vs pointing action & Collectors (=~curators) \\
\ds{MoteStrain} & Which source stream: humidity vs temperature & Unsupervised stream correlation tracking; no class target & Curators \\
\ds{FreezerRegularTrain} & Which of two freezers in one house & REFIT load monitoring; no class target & Curators \\
\ds{FreezerSmallTrain} & As FreezerRegularTrain; smaller training split & As FreezerRegularTrain & Curators \\
\ds{FaceAll}$^\dagger$ & Which of 14 graduate students, from head profile & Head profiles recorded at UCR; no applied problem & Collectors (=~curators) \\
\ds{FacesUCR}$^\dagger$ & As FaceAll; rotation-aligned, different split & As FaceAll & Collectors (=~curators) \\
\ds{Haptics} & Which of five users entered the passgraph & None documented; passgraph entry on a touchscreen & Unknown \\
\end{longtable}

%% file: tables/labels_replacement.tex
\begin{longtable}{@{}P{3.2cm} P{3.2cm} P{4.4cm} P{1.6cm}@{}}
\caption{Task-construction cases in the current audit: replacement. None means no classification target is documented, not that the source data had no scientific purpose.}\label{tab:labels-replacement} \\
\toprule
Dataset & Benchmark target & Source target or purpose & Label by \\
\midrule
\endfirsthead
\multicolumn{4}{l}{\emph{(continued)}} \\
\toprule
Dataset & Benchmark target & Source target or purpose & Label by \\
\midrule
\endhead
\midrule
\multicolumn{4}{r}{\emph{continues}} \\
\endfoot
\bottomrule
\endlastfoot
\ds{AllGestureWiimoteX} & Which of ten gestures & User identity; gesture was a protocol factor & Curators \\
\ds{AllGestureWiimoteY} & As AllGestureWiimoteX & As AllGestureWiimoteX & Curators \\
\ds{AllGestureWiimoteZ} & As AllGestureWiimoteX & As AllGestureWiimoteX & Curators \\
\ds{Earthquakes} & Whether a large event follows a window; curator thresholds & Continuous seismic monitoring; no class target & Curators \\
\ds{HouseTwenty} & Whole-house aggregate vs summed dryer/washer channels & REFIT load monitoring; no class target & Curators \\
\ds{ECG5000} & Five automatically assigned beat types & No clinician beat labels used for this record & Curators \\
\ds{ToeSegmentation1} & Normal vs acted abnormal gait & CMU motion descriptions; no defined task & Curators \\
\ds{ToeSegmentation2} & As ToeSegmentation1 & As ToeSegmentation1 & Curators \\
\ds{ChlorineConcentration} & Three undocumented classes & None; EPANET simulation built to test missing-value imputation & Curators \\
\ds{BeetleFly} & Beetle vs fly, a pair carved out of 70 classes & Shape retrieval over 70 MPEG-7 classes; no applied problem & Curators \\
\ds{BirdChicken} & Bird vs chicken, a pair carved out of 70 classes & As BeetleFly & Curators \\
\ds{FiftyWords} & Which of 50 word types & Word spotting by word identity in historical manuscripts & Curators \\
\ds{WordSynonyms} & Which of 25 curator-defined word groups & As FiftyWords & Curators \\
\end{longtable}

%% file: tables/labels_toy.tex
\begin{longtable}{@{}P{3.2cm} P{3.2cm} P{4.1cm} P{1.9cm}@{}}
\caption{Task-construction cases in the current audit: toy source task. None means no classification target is documented, not that the source data had no scientific purpose. $^\dagger$ also listed under another mechanism.}\label{tab:labels-toy} \\
\toprule
Dataset & Benchmark target & Source target or purpose & Label by \\
\midrule
\endfirsthead
\multicolumn{4}{l}{\emph{(continued)}} \\
\toprule
Dataset & Benchmark target & Source target or purpose & Label by \\
\midrule
\endhead
\midrule
\multicolumn{4}{r}{\emph{continues}} \\
\endfoot
\bottomrule
\endlastfoot
\ds{BME} & Bell at begin, none, or at end of a central plate & None; unpublished generator (Grenoble LIG/AMA) & Generator \\
\ds{UMD} & Bell pointing up, absent, or pointing down & As BME & Generator \\
\ds{CBF} & Cylinder, bell or funnel shape in noise & None; demo signal for local feature extraction & Generator \\
\ds{SyntheticControl} & Six control-chart patterns & None; no process measured, generated to test a similarity method & Generator \\
\ds{TwoPatterns} & Ordered pair of up/down steps in noise & None; designed for thesis experiments & Generator \\
\ds{Mallat} & Eight generated signal variants; classes undocumented & None; wavelet test signal & Generator \\
\ds{ShapeletSim} & Which of two shapes is embedded in noise & None; simulator built so that shapelet methods win & Generator \\
\ds{SmoothSubspace} & Which five-step window carries the signal & Cluster identity, to test a $k$-means variant & Generator \\
\ds{Trace} & Four feature/fault pairs & Sixteen simulated plant transient/failure classes & Generator \\
\ds{GunPoint} & Gun-draw vs pointing & Staged at UCR as a similarity-search benchmark & Collectors (=~curators) \\
\ds{GunPointAgeSpan} & Gun-draw vs pointing, pooled over actors and years & GunPoint re-staged for the 2018 archive release & Collectors (=~curators) \\
\ds{GunPointMaleVersusFemale}$^\dagger$ & Actor sex; confounded with identity (two actors) & As GunPointAgeSpan & Collectors (=~curators) \\
\ds{GunPointOldVersusYoung}$^\dagger$ & Recording session: 2003 vs 2018 & As GunPointAgeSpan & Collectors (=~curators) \\
\ds{Yoga}$^\dagger$ & Which of two actors, from silhouettes & Actor discrimination, posed by the collectors as a demo task & Collectors (=~curators) \\
\ds{FaceAll}$^\dagger$ & Which of 14 graduate students, from head profile & Head profiles recorded at UCR; no applied problem & Collectors (=~curators) \\
\ds{FacesUCR}$^\dagger$ & As FaceAll; rotation-aligned, different split & As FaceAll & Collectors (=~curators) \\
\ds{Symbols} & Six classes; three symbols documented & Copying experiment with 13 participants; no publication & Collectors (=~curators) \\
\end{longtable}

%% file: tables/duplicates.tex
\begin{longtable}{@{}P{3.2cm} l r r r r r@{}}
\caption{Exact cross-split matches and MiniRocket accuracy, retained from the original generated results. Counts refer to test series, not pairs. Mixed counts duplicate groups with conflicting labels; -- denotes none.}\label{tab:duplicates} \\
\toprule
Dataset & Archive & Test = train & \% & Acc. dup. & Acc. rest & Mixed \\
\midrule
\endfirsthead
\multicolumn{7}{l}{\emph{(continued)}} \\
\toprule
Dataset & Archive & Test = train & \% & Acc. dup. & Acc. rest & Mixed \\
\midrule
\endhead
\midrule
\multicolumn{7}{r}{\emph{continues}} \\
\endfoot
\bottomrule
\endlastfoot
\ds{PEMS-SF} & UEA & 79 / 173 & 45.7 & 1.000 & 0.681 & --  \\
\ds{PowerCons} & UCR & 67 / 180 & 37.2 & 1.000 & 0.991 & --  \\
\ds{Crop} & UCR & 2,753 / 16,800 & 16.4 & 0.992 & 0.722 & --  \\
\ds{WormsTwoClass} & UCR & 10 / 77 & 13.0 & 1.000 & 0.746 & --  \\
\ds{Worms} & UCR & 10 / 77 & 13.0 & 1.000 & 0.701 & --  \\
\ds{Libras} & UEA & 19 / 180 & 10.6 & 1.000 & 0.907 & --  \\
\ds{InsectEPGSmallTrain} & UCR & 15 / 249 & 6.0 & 1.000 & 0.991 & --  \\
\ds{EigenWorms} & UEA & 7 / 131 & 5.3 & 1.000 & 0.935 & --  \\
\ds{ElectricDevices} & UCR & 208 / 7,711 & 2.7 & 0.365 & 0.752 & 54  \\
\ds{ToeSegmentation1} & UCR & 4 / 228 & 1.8 & 1.000 & 0.960 & --  \\
\ds{InlineSkate} & UCR & 6 / 550 & 1.1 & 1.000 & 0.471 & --  \\
\ds{FreezerSmallTrain} & UCR & 27 / 2,850 & 0.9 & 1.000 & 0.967 & --  \\
\ds{PLAID} & UCR & 5 / 537 & 0.9 & 1.000 & 0.930 & --  \\
\ds{ShapesAll} & UCR & 2 / 600 & 0.3 & 1.000 & 0.923 & --  \\
\ds{SwedishLeaf} & UCR & 2 / 625 & 0.3 & 1.000 & 0.961 & --  \\
\ds{Strawberry} & UCR & 1 / 370 & 0.3 & 0.000 & 0.986 & 1  \\
\ds{MedicalImages} & UCR & 1 / 760 & 0.1 & 1.000 & 0.791 & --  \\
\ds{FaceAll} & UCR & 1 / 1,690 & 0.1 & 1.000 & 0.806 & --  \\
\ds{MixedShapesRegularTrain} & UCR & 1 / 2,425 & 0.0 & 1.000 & 0.971 & --  \\
\ds{FreezerRegularTrain} & UCR & 1 / 2,850 & 0.0 & 1.000 & 1.000 & --  \\
\end{longtable}

%% file: tables/split_audit.tex
\begingroup
\setlength{\LTleft}{0pt}
\setlength{\LTright}{0pt}
\setlength{\tabcolsep}{3pt}
\renewcommand{\arraystretch}{1.12}
\newcommand{\SplitAuditName}[1]{\textit{\hspace{0pt}#1}}
\begin{longtable}{@{}>{\raggedright\arraybackslash}p{0.22\linewidth} >{\raggedright\arraybackslash}p{0.15\linewidth} >{\raggedright\arraybackslash}p{0.115\linewidth} >{\raggedright\arraybackslash}p{\dimexpr0.515\linewidth-6\tabcolsep\relax}@{}}
\caption{UCR split-audit categorization (128 datasets): Concern 32.03\%; Unclear 50.78\%; Separated 5.47\%; Related periods 5.47\%; Generated 6.25\%.}\label{tab:splits} \\
\toprule
Dataset & Relevant new case & Verdict & What the split tests and why a concern can rise \\
\midrule
\endfirsthead
\multicolumn{4}{l}{\emph{UCR split audit (continued)}} \\
\toprule
Dataset & Relevant new case & Verdict & What the split tests and why a concern remains \\
\midrule
\endhead
\midrule
\multicolumn{4}{r}{\emph{continues on the next page}} \\
\endfoot
\bottomrule
\endlastfoot
\hypertarget{split-audit:ACSF1}{}\SplitAuditName{ACSF1} & New appliances & Unclear & The 100/100 split matches the source's Test Protocol 1.0, which trains on the first measurement session and tests on the second session of the same appliances. The archive does not say which protocol was used, and the series length does not match the documented one-hour sessions. Whether the test presents familiar appliances again or appliances the model has never seen therefore cannot be established. A concern arises only if the result is presented as performance on new appliances. \\
\hypertarget{split-audit:Adiac}{}\SplitAuditName{Adiac} & New independently collected samples & Unclear & The cases are outlines of tiny biological specimens. We do not know whether samples grown from the same biological source, or photographed on the same microscope slide, occur in both sets. They could share growth or imaging conditions. The records do not let us check whether the test samples were collected independently. \\
\hypertarget{split-audit:AllGestureWiimoteX}{}\SplitAuditName{AllGestureWiimoteX} & New people & Unclear & The data contain gestures from ten people. The archive shuffled and sampled recordings, but does not say that all recordings from a person were kept in one set. We therefore do not know whether test users are new. This matters if the intended application is to recognise gestures from people the model has never seen. \\
\hypertarget{split-audit:AllGestureWiimoteY}{}\SplitAuditName{AllGestureWiimoteY} & New people & Unclear & Same as \hyperlink{split-audit:AllGestureWiimoteX}{\SplitAuditName{AllGestureWiimoteX}}. \\
\hypertarget{split-audit:AllGestureWiimoteZ}{}\SplitAuditName{AllGestureWiimoteZ} & New people & Unclear & Same as \hyperlink{split-audit:AllGestureWiimoteX}{\SplitAuditName{AllGestureWiimoteX}}. \\
\hypertarget{split-audit:ArrowHead}{}\SplitAuditName{ArrowHead} & New physical objects & Unclear & Each case is an outline of an arrowhead. The source assigned outlines to a 36/175 split at random. We cannot check whether multiple views or representations of the same object were kept together. \\
\hypertarget{split-audit:BME}{}\SplitAuditName{BME} & New generated examples & Generated & This is a generated task rather than measurements from people, devices or specimens. The corresponding source-separation question is therefore not assessed here. Its split would instead need to be examined through the generation procedure; synthetic origin alone is not evidence of leakage. \\
\hypertarget{split-audit:Beef}{}\SplitAuditName{Beef} & New physical samples & Unclear & The data are spectral measurements of beef samples. It is not documented whether repeated measurements of one physical sample were kept in the same set. Without that information, performance on new samples cannot be distinguished from performance on further measurements of familiar samples. \\
\hypertarget{split-audit:BeetleFly}{}\SplitAuditName{BeetleFly} & New source images & Unclear & The cases are outlines extracted from images of beetles and flies. We do not know how the original images were assigned to training and testing. If different versions of one image occur in both sets, the model could recognise a familiar shape. Such shared images have not been verified. \\
\hypertarget{split-audit:BirdChicken}{}\SplitAuditName{BirdChicken} & New source images & Unclear & Same as \hyperlink{split-audit:BeetleFly}{\SplitAuditName{BeetleFly}}. \\
\hypertarget{split-audit:CBF}{}\SplitAuditName{CBF} & New generated examples & Generated & This is a generated task rather than measurements from people, devices or specimens. The corresponding source-separation question is therefore not assessed here. Its split would instead need to be examined through the generation procedure; synthetic origin alone is not evidence of leakage. \\
\hypertarget{split-audit:Car}{}\SplitAuditName{Car} & New vehicles or recordings & Unclear & The outlines were extracted from traffic videos. Several frames can show the same vehicle under very similar conditions. The documentation does not establish whether all frames of a vehicle or video were kept in one set, so performance on new vehicles or capture occasions cannot be checked. \\
\hypertarget{split-audit:Chinatown}{}\SplitAuditName{Chinatown} & New days or later periods & Related periods & A test day may be close to a training day and share its season or local events. These are different days, so this is not automatically leakage. No source says whether the split holds out a later period that follows all the training days. \\
\hypertarget{split-audit:ChlorineConcentration}{}\SplitAuditName{ChlorineConcentration} & New simulation runs & Concern & The audit describes cases taken at nearby times in one simulated system. A training case and test case can therefore describe almost the same system conditions. The result tests more cases from that simulation. It does not show whether the model works on a new simulation run; similarity between cases would need a separate check. \\
\hypertarget{split-audit:CinCECGTorso}{}\SplitAuditName{CinCECGTorso} & New recording occasions & Concern & The task is to identify four people from their ECG measurements. The archive defines each class as one person, so training and test share those people, and very likely one recording each. The same identities are expected in an identification task, but new recordings are not tested. The model may recognise characteristics of the recording as well as the person. \\
\hypertarget{split-audit:Coffee}{}\SplitAuditName{Coffee} & New physical samples & Unclear & The cases are spectral measurements of coffee. We do not know whether a sample was measured more than once and whether those measurements were kept together. If one measurement is in training and another in testing, the test sample is familiar. The missing sample information prevents us from checking this possibility. \\
\hypertarget{split-audit:Computers}{}\SplitAuditName{Computers} & New appliances or households & Unclear & The source measured each physical appliance over many days. We do not know whether all days from an appliance or household were kept in one set. If the same appliance appears in both sets, the model may recognise its usual pattern. That would be a narrower test than recognising appliance types in new homes. \\
\hypertarget{split-audit:CricketX}{}\SplitAuditName{CricketX} & New people & Unclear & The data contain repeated gestures, but the files do not say which person produced each case. We cannot check whether all gestures from a person were kept in one set. A test gesture might come from someone already represented in training. Performance on people the model has never seen is therefore not established. \\
\hypertarget{split-audit:CricketY}{}\SplitAuditName{CricketY} & New people & Unclear & Same as \hyperlink{split-audit:CricketX}{\SplitAuditName{CricketX}}. \\
\hypertarget{split-audit:CricketZ}{}\SplitAuditName{CricketZ} & New people & Unclear & Same as \hyperlink{split-audit:CricketX}{\SplitAuditName{CricketX}}. \\
\hypertarget{split-audit:Crop}{}\SplitAuditName{Crop} & New fields, scenes or seasons & Concern & The cases are pixel sequences from one scene and season. Nearby pixels may describe the same field and growing conditions. The value check also finds 2,753 of 16,800 test cases identical to training cases (Table~\ref{tab:duplicates}). This makes the test different from evaluating new fields, scenes or seasons. \\
\hypertarget{split-audit:DiatomSizeReduction}{}\SplitAuditName{DiatomSizeReduction} & New biological sources & Unclear & The sources explain size variation within each class by successive generations of a clone, but do not say how the split was drawn or whether training and test come from the same clone. If they do, a new specimen from a familiar line is a narrower test than a specimen from an independently collected source. The intended application should make clear which of these is relevant. \\
\hypertarget{split-audit:DistalPhalanxOutlineAgeGroup}{}\SplitAuditName{DistalPhalanxOutlineAgeGroup} & New patients & Unclear & The cases are finger-bone outlines extracted from X-rays. We do not know whether one patient contributed several images or whether those images were kept together. If one patient appears in both sets, the model may benefit from familiar anatomy. Patient identifiers are needed to check performance on new patients. \\
\hypertarget{split-audit:DistalPhalanxOutlineCorrect}{}\SplitAuditName{DistalPhalanxOutlineCorrect} & New patients & Unclear & Same as \hyperlink{split-audit:DistalPhalanxOutlineAgeGroup}{\SplitAuditName{DistalPhalanxOutlineAgeGroup}}. \\
\hypertarget{split-audit:DistalPhalanxTW}{}\SplitAuditName{DistalPhalanxTW} & New patients & Unclear & Same as \hyperlink{split-audit:DistalPhalanxOutlineAgeGroup}{\SplitAuditName{DistalPhalanxOutlineAgeGroup}}. \\
\hypertarget{split-audit:DodgerLoopDay}{}\SplitAuditName{DodgerLoopDay} & New days or later periods & Related periods & Days from one traffic detector were assigned at random. Training and test days can be part of the same baseball event or share similar traffic conditions. The test contains different days, but does not specifically evaluate a later period or another detector. This relationship should be considered when interpreting the score. \\
\hypertarget{split-audit:DodgerLoopGame}{}\SplitAuditName{DodgerLoopGame} & New days or later periods & Related periods & Same as \hyperlink{split-audit:DodgerLoopDay}{\SplitAuditName{DodgerLoopDay}}. \\
\hypertarget{split-audit:DodgerLoopWeekend}{}\SplitAuditName{DodgerLoopWeekend} & New days or later periods & Related periods & Same as \hyperlink{split-audit:DodgerLoopDay}{\SplitAuditName{DodgerLoopDay}}. \\
\hypertarget{split-audit:ECG200}{}\SplitAuditName{ECG200} & New recording sessions of a known patient & Concern & Each case is a heartbeat. The source thesis drew all 200 beats from one Holter recording of a single patient, so training and test beats share the patient and the recording session. For patient-specific monitoring a new case is a new recording session of that patient, which the split cannot test. \\
\hypertarget{split-audit:ECG5000}{}\SplitAuditName{ECG5000} & New patients or recording occasions & Concern & The source record identified by the audit contains heartbeats from one patient during one long recording. Training and test beats therefore share that patient and recording. This can test more beats from that record, but it cannot establish performance on another patient or after the patient is recorded again. \\
\hypertarget{split-audit:ECGFiveDays}{}\SplitAuditName{ECGFiveDays} & Independent recordings & Concern & The label says which of two recording dates produced a heartbeat. Training and test beats come from those same two recordings of one person. The model may distinguish conditions specific to those recordings. The test does not show that the distinction holds when further independent recordings are collected. \\
\hypertarget{split-audit:EOGHorizontalSignal}{}\SplitAuditName{EOGHorizontalSignal} & New people or recording occasions & Unclear & The eye-movement samples come from six people. The audit describes splitting individual movements rather than keeping all movements from a person together. We cannot establish performance on unfamiliar users. Sharing users can suit a personalised system, but its test should explain whether they were measured again on a separate occasion. \\
\hypertarget{split-audit:EOGVerticalSignal}{}\SplitAuditName{EOGVerticalSignal} & New people or sessions & Unclear & Same as \hyperlink{split-audit:EOGHorizontalSignal}{\SplitAuditName{EOGHorizontalSignal}}. \\
\hypertarget{split-audit:Earthquakes}{}\SplitAuditName{Earthquakes} & New periods or event groups & Related periods & The time windows are described as non-overlapping, so shared time samples are not an established problem here. However, windows from one earthquake sequence can be assigned to both sets. They may share the same broader event. The split does not specifically test later periods or separately collected event groups. \\
\hypertarget{split-audit:ElectricDevices}{}\SplitAuditName{ElectricDevices} & New appliances or households & Concern & The source kept each household in one set, but the shipped files differ in size from the published split, so it is unverified that the archive split keeps this. The value check also finds 208 test cases identical to training cases (Table~\ref{tab:duplicates}). Also, 54 of its repeated-value groups carry conflicting labels. Most matches are two-valued on/off days, so they need not mean shared households. \\
\hypertarget{split-audit:EthanolLevel}{}\SplitAuditName{EthanolLevel} & New bottle types & Separated & The archive paper says that a bottle type is never used in both training and testing \citep{dau2019ucr}. The test therefore checks measurements involving unfamiliar bottle types. This is a documented source separation. It does not, by itself, establish that every other recording condition is different between the sets. \\
\hypertarget{split-audit:FaceAll}{}\SplitAuditName{FaceAll} & New photography occasions & Concern & The task identifies people, so those identities are expected in both sets. The concern is whether test photographs were taken on a different occasion. The audit does not establish this. Similar photos taken under the same conditions may make identification easier than recognising the person in newly collected photographs. The value check also reports one test case identical to training data (Table~\ref{tab:duplicates}). \\
\hypertarget{split-audit:FaceFour}{}\SplitAuditName{FaceFour} & New source images or sessions & Unclear & The available record does not adequately describe the people, source photographs or capture sessions. We cannot check whether related images or acquisitions occur in both sets. No specific overlap is established by this absence of documentation. \\
\hypertarget{split-audit:FacesUCR}{}\SplitAuditName{FacesUCR} & New photography occasions & Unclear & The task uses reprocessed photographs from the same collection as FaceAll. Identifying familiar people is expected, but the audit does not establish that test photos were taken on a separate occasion. Similar lighting, pose or camera conditions could help recognition without showing that it works in newly collected photographs. \\
\hypertarget{split-audit:FiftyWords}{}\SplitAuditName{FiftyWords} & New manuscript pages & Unclear & The word images come from one writer's manuscripts. Words on the same page can share handwriting and scanning conditions. We do not know whether whole pages were kept in one set. The test may therefore contain more words from familiar pages, rather than words from pages the model has never seen. \\
\hypertarget{split-audit:Fish}{}\SplitAuditName{Fish} & New physical specimens & Unclear & The cases are outlines derived from fish images. The description does not establish how physical specimens and any repeated views were assigned to the two sets. New-specimen evaluation therefore cannot be verified; repeated views are a possibility to check, not an observed overlap. \\
\hypertarget{split-audit:FordA}{}\SplitAuditName{FordA} & New vehicles or recording occasions & Unclear & The files do not identify the vehicle, test rig or recording occasion behind each case. We therefore cannot check whether these sources were kept apart. The split may be appropriate, but the available information does not show that testing uses vehicles or recording occasions absent from training. \\
\hypertarget{split-audit:FordB}{}\SplitAuditName{FordB} & New conditions and recordings & Unclear & The audit describes test data collected under different, noisier operating conditions. This tests a change in conditions; it does not demonstrate shared recordings. We still do not know which vehicles or recording occasions produced the cases, so we cannot check whether those sources are different between the sets. \\
\hypertarget{split-audit:FreezerRegularTrain}{}\SplitAuditName{FreezerRegularTrain} & New physical freezers & Concern & The source contains ongoing measurements of two freezers in one house. Pieces of the same recording can share appliance settings and operating conditions. The audit does not show that testing was kept for a later period or a new recording occasion. The value check separately reports 1 of 2,850 test cases matching training data. \\
\hypertarget{split-audit:FreezerSmallTrain}{}\SplitAuditName{FreezerSmallTrain} & New physical freezers & Concern & Same as \hyperlink{split-audit:FreezerRegularTrain}{\SplitAuditName{FreezerRegularTrain}}. This smaller-training-set variant has 27 of 2,850 test cases matching training data in the value check. \\
\hypertarget{split-audit:Fungi}{}\SplitAuditName{Fungi} & New biological strains & Concern & The source measured fungal strains repeatedly. The audit reports that testing includes another measurement of a strain used in training. A model could recognise that familiar strain. This is a narrower test than identifying the species of a newly collected strain, for which all measurements of a strain should be kept together. \\
\hypertarget{split-audit:GestureMidAirD1}{}\SplitAuditName{GestureMidAirD1} & New people & Separated & The archive description says that no person occurs in both sets \citep{tscarchive_descriptions}. The test therefore uses people absent from training. There is disagreement in the records about which participant group is training, but both describe keeping the groups separate. No shared-person concern is identified by that description. \\
\hypertarget{split-audit:GestureMidAirD2}{}\SplitAuditName{GestureMidAirD2} & New people & Separated & Same as \hyperlink{split-audit:GestureMidAirD1}{\SplitAuditName{GestureMidAirD1}}. \\
\hypertarget{split-audit:GestureMidAirD3}{}\SplitAuditName{GestureMidAirD3} & New people & Separated & Same as \hyperlink{split-audit:GestureMidAirD1}{\SplitAuditName{GestureMidAirD1}}. \\
\hypertarget{split-audit:GesturePebbleZ1}{}\SplitAuditName{GesturePebbleZ1} & New people & Concern & The same four users performed gestures on two occasions. Training uses the first occasion and testing the second \citep{dau2019ucr}. Every test user is also in training, so the score can rest on per-user motion signatures and does not measure recognition for new wearers of the watch; \hyperlink{split-audit:GesturePebbleZ2}{\SplitAuditName{GesturePebbleZ2}} is the subject-disjoint version. \\
\hypertarget{split-audit:GesturePebbleZ2}{}\SplitAuditName{GesturePebbleZ2} & New people & Separated & Training uses two people and testing the other two \citep{dau2019ucr}. The model therefore sees unfamiliar users in the test. This differs from measuring familiar users again. The description establishes separation of people, although it does not establish separation of every other possible recording condition. \\
\hypertarget{split-audit:GunPoint}{}\SplitAuditName{GunPoint} & New people or recording occasions & Concern & All 200 series come from one recording session of two actors, so training and test necessarily share that session and the split cannot test newly collected recordings. Which actor is on which side is not recorded. Having only two actors also limits how broadly results can be interpreted. \\
\hypertarget{split-audit:GunPointAgeSpan}{}\SplitAuditName{GunPointAgeSpan} & New people or recording occasions & Concern & The archive pools actions from two actors recorded in two years \citep{dau2019ucr}. A check of the shipped files finds every actor and year combination in both sets, so the same two people and recording sessions appear on both sides. The score therefore concerns more actions from that collection. It does not establish performance on unfamiliar people or measurements collected on a new occasion. \\
\hypertarget{split-audit:GunPointMaleVersusFemale}{}\SplitAuditName{GunPointMaleVersusFemale} & New people or recording occasions & Concern & The two classes distinguish the two actors, and recordings from both years are pooled \citep{dau2019ucr}. A model can recognise these particular people to predict the label. The audit reports no new recording occasion in testing. The result also does not establish the same distinction for other people. \\
\hypertarget{split-audit:GunPointOldVersusYoung}{}\SplitAuditName{GunPointOldVersusYoung} & New recordings for each age condition & Concern & The label distinguishes the original recording years \citep{dau2019ucr}. Both sets use actions recorded on those occasions. Differences in equipment or recording conditions could help the model along with age-related changes. Further recordings under each age condition would be needed to check whether the result reflects age rather than those occasions. \\
\hypertarget{split-audit:Ham}{}\SplitAuditName{Ham} & New physical samples & Unclear & Each physical ham supplies several image measurements. We do not know whether all measurements of a ham were kept in one set. The test could therefore contain another measurement of a familiar ham. Not finding very similar pairs in the original check does not prove that different hams were used. \\
\hypertarget{split-audit:HandOutlines}{}\SplitAuditName{HandOutlines} & New patients & Unclear & The outlines come from hand X-rays. We do not know whether one patient contributed several images or whether those images were kept together. The test could contain a new image of a familiar patient. Patient identifiers or an explicit patient grouping rule are needed to check this. \\
\hypertarget{split-audit:Haptics}{}\SplitAuditName{Haptics} & New recording occasions & Unclear & The labels identify users from repeated measurements. We do not know whether the test measurements were collected on occasions separate from training. Familiar identities are expected, but the model might also recognise conditions specific to one recording. Identification when a user is measured again remains unverified. \\
\hypertarget{split-audit:Herring}{}\SplitAuditName{Herring} & New physical specimens & Unclear & The cases are outlines derived from fish ear-bone images. The description does not establish how physical specimens and any repeated views were assigned to the two sets. New-specimen evaluation therefore cannot be verified; repeated views are a possibility to check, not an observed overlap. \\
\hypertarget{split-audit:HouseTwenty}{}\SplitAuditName{HouseTwenty} & New households & Concern & The cases are pieces of one household's ongoing recording, so training and testing share the same home, its appliances and its routine. The test measures recognition within that home; performance in another home, which load monitoring needs, is not tested. \\
\hypertarget{split-audit:InlineSkate}{}\SplitAuditName{InlineSkate} & New recording occasions & Concern & Six of the 550 test series are value-identical to training series (Table~\ref{tab:duplicates}). If the classes are individuals, training and test cycles may also come from the same treadmill run and share its person, equipment and conditions, but no source confirms this. The test does not show whether recognition works after collecting another recording. This interpretation also needs confirmation of what the original class labels identify. \\
\hypertarget{split-audit:InsectEPGRegularTrain}{}\SplitAuditName{InsectEPGRegularTrain} & New insects or recordings & Unclear & The cases are pieces of long insect recordings. No insect or recording identifier ships with the data, so we cannot check whether pieces of one insect were kept in one set. Pieces in both sets could share the plant, sensor setup and nearby feeding behaviour, but this is not established. Unlike \hyperlink{split-audit:InsectEPGSmallTrain}{\SplitAuditName{InsectEPGSmallTrain}}, which shares its test set, this training set has no exact matches with the test data. \\
\hypertarget{split-audit:InsectEPGSmallTrain}{}\SplitAuditName{InsectEPGSmallTrain} & New insects or recordings & Concern & The cases are pieces of long insect recordings. No insect or recording identifier ships with the data, so whether pieces of one insect occur in both sets is not established. The value check finds 15 of the 249 test series identical to series in this small training set (Table~\ref{tab:duplicates}), so a model can score them by recall. \\
\hypertarget{split-audit:InsectWingbeatSound}{}\SplitAuditName{InsectWingbeatSound} & New insect groups or recordings & Unclear & The source recorded insects in groups inside cages. Cases from one group can share its insects and recording conditions. The files do not identify those groups, so we cannot check whether a whole group was kept in one set. Shared groups are possible, but have not been verified. \\
\hypertarget{split-audit:ItalyPowerDemand}{}\SplitAuditName{ItalyPowerDemand} & New days or later periods & Related periods & Training and test days can share the same season and similar weather. Different days are tested, so this is not automatically leakage. However, the split does not specifically test a later period or another grid. \\
\hypertarget{split-audit:LargeKitchenAppliances}{}\SplitAuditName{LargeKitchenAppliances} & New appliances or households & Unclear & Same as \hyperlink{split-audit:Computers}{\SplitAuditName{Computers}}. \\
\hypertarget{split-audit:Lightning2}{}\SplitAuditName{Lightning2} & New recorded events & Unclear & Each series is one recorded lightning event, but no source documents how the 121 records were divided or whether events of one flash or storm sit on both sides. \\
\hypertarget{split-audit:Lightning7}{}\SplitAuditName{Lightning7} & New recorded events & Unclear & Each series is one recorded lightning event. The source evaluated by random cross-validation over all 143 records \citep{eads2002zeus}; who built the fixed 70/73 split, and whether strokes of one flash sit on both sides, is not documented. \\
\hypertarget{split-audit:Mallat}{}\SplitAuditName{Mallat} & New generated examples & Generated & This is a generated task rather than measurements from people, devices or specimens. The corresponding source-separation question is therefore not assessed here. Its split would instead need to be examined through the generation procedure; synthetic origin alone is not evidence of leakage. \\
\hypertarget{split-audit:Meat}{}\SplitAuditName{Meat} & New physical samples & Unclear & The source measured each of 60 physical samples twice. We do not know whether the two measurements of a sample were kept together. If one goes into training and the other into testing, the model sees another measurement of a familiar sample. The sample assignment is needed to check whether this happened. \\
\hypertarget{split-audit:MedicalImages}{}\SplitAuditName{MedicalImages} & New patients or images & Concern & The files do not identify the patient or original image behind each series. Different series might therefore come from one patient or image. We cannot check whether those sources were kept apart. Missing identifiers leave this question unanswered; they do not prove that sources were shared. Separately, the value check reports one test case identical to training data (Table~\ref{tab:duplicates}). \\
\hypertarget{split-audit:MelbournePedestrian}{}\SplitAuditName{MelbournePedestrian} & New days or later periods & Related periods & Training and test days come from the same ten city sensors and the same year 2017; the assignment rule is undocumented. Training and test days can share local events, seasons and familiar sensors. The test checks more days from that collection. It does not specifically check a later period or a sensor that was absent from training. \\
\hypertarget{split-audit:MiddlePhalanxOutlineAgeGroup}{}\SplitAuditName{MiddlePhalanxOutlineAgeGroup} & New patients & Unclear & Same as \hyperlink{split-audit:DistalPhalanxOutlineAgeGroup}{\SplitAuditName{DistalPhalanxOutlineAgeGroup}}. \\
\hypertarget{split-audit:MiddlePhalanxOutlineCorrect}{}\SplitAuditName{MiddlePhalanxOutlineCorrect} & New patients & Unclear & Same as \hyperlink{split-audit:DistalPhalanxOutlineAgeGroup}{\SplitAuditName{DistalPhalanxOutlineAgeGroup}}. \\
\hypertarget{split-audit:MiddlePhalanxTW}{}\SplitAuditName{MiddlePhalanxTW} & New patients & Unclear & Same as \hyperlink{split-audit:DistalPhalanxOutlineAgeGroup}{\SplitAuditName{DistalPhalanxOutlineAgeGroup}}. \\
\hypertarget{split-audit:MixedShapesRegularTrain}{}\SplitAuditName{MixedShapesRegularTrain} & New images or physical objects & Concern & The cases are outlines from shape images. We do not know whether several views or processed versions of one original object can occur in both sets. If they do, the model may recognise a familiar shape. The source-image information needed to check this is not described in the audit. Separately, the value check reports one test case identical to training data (Table~\ref{tab:duplicates}). \\
\hypertarget{split-audit:MixedShapesSmallTrain}{}\SplitAuditName{MixedShapesSmallTrain} & New source images or objects & Unclear & Same uncertainty about the original images as \hyperlink{split-audit:MixedShapesRegularTrain}{\SplitAuditName{MixedShapesRegularTrain}}. The one value match mentioned there concerns MixedShapesRegularTrain, not this variant. \\
\hypertarget{split-audit:MoteStrain}{}\SplitAuditName{MoteStrain} & Independent recordings & Concern & The audit traces cases to nearby pieces of a small number of sensor recordings, with the recording defining the class. Training and testing therefore share those original recordings. The model may recognise their particular conditions. The test does not show that the distinction works across recordings collected independently. \\
\hypertarget{split-audit:NonInvasiveFetalECGThorax1}{}\SplitAuditName{NonInvasiveFetalECGThorax1} & Independent recordings & Unclear & The original audit thinks the labels identify recordings and that training and test heartbeats come from those same recordings. However, the source mapping is not confirmed. If correct, testing checks more beats from familiar recordings. This remains a possible concern until the original sources and labels are confirmed. \\
\hypertarget{split-audit:NonInvasiveFetalECGThorax2}{}\SplitAuditName{NonInvasiveFetalECGThorax2} & Independent source recordings & Unclear & Same as \hyperlink{split-audit:NonInvasiveFetalECGThorax1}{\SplitAuditName{NonInvasiveFetalECGThorax1}}. \\
\hypertarget{split-audit:OSULeaf}{}\SplitAuditName{OSULeaf} & New physical specimens & Unclear & The cases are outlines derived from leaf images. The description does not establish how physical specimens and any repeated views were assigned to the two sets. New-specimen evaluation therefore cannot be verified; repeated views are a possibility to check, not an observed overlap. \\
\hypertarget{split-audit:OliveOil}{}\SplitAuditName{OliveOil} & New physical samples & Unclear & The cases are spectral measurements of oils. The source measured 60 oils in duplicate; the archive ships 60 spectra and does not say whether they are one per oil or both replicates of 30 oils. Looking for similar sequences cannot tell us which sample produced them. The test's use of genuinely new samples therefore remains unverified. \\
\hypertarget{split-audit:PLAID}{}\SplitAuditName{PLAID} & New physical appliances & Concern & The source recorded individual appliances several times. We do not know whether all recordings of an appliance were kept in one set. Familiar appliances might occur in testing, but this is not established by the missing information. Separately, the value check finds five test cases identical to training cases. \\
\hypertarget{split-audit:PhalangesOutlinesCorrect}{}\SplitAuditName{PhalangesOutlinesCorrect} & New X-rays or patients & Unclear & Several bone outlines can come from one hand X-ray. Unless kept together, one image can supply outlines to both sets, letting the model benefit from familiar image conditions. The audit does not describe such grouping. Image or patient identifiers are needed to establish whether this possible overlap actually occurs. \\
\hypertarget{split-audit:Phoneme}{}\SplitAuditName{Phoneme} & New speakers or spoken recordings & Unclear & One spoken word can supply several phoneme segments. We do not know whether complete word recordings or speakers were kept in one set. Segments in both sets could share a voice and recording conditions. The test's use of new speakers or separately recorded words remains unverified. \\
\hypertarget{split-audit:PickupGestureWiimoteZ}{}\SplitAuditName{PickupGestureWiimoteZ} & New recording occasions & Unclear & The labels identify people, with several gesture repetitions per person. We do not know whether those repetitions were recorded on separate occasions or how the occasions were assigned to the sets. The test may check more repetitions from a familiar recording. Identification after recording the person again is not established. \\
\hypertarget{split-audit:PigAirwayPressure}{}\SplitAuditName{PigAirwayPressure} & New recording occasions & Concern & The task is to identify pigs. Training uses the first pieces of each pig's before/after pressure recordings; testing uses later pieces of the same recordings \citep{dau2019ucr}. The same pigs are expected in an identification task. The concern is that the model is tested on familiar recordings, rather than after measuring those pigs again. \\
\hypertarget{split-audit:PigArtPressure}{}\SplitAuditName{PigArtPressure} & New sessions of known animals & Concern & Same as \hyperlink{split-audit:PigAirwayPressure}{\SplitAuditName{PigAirwayPressure}}. \\
\hypertarget{split-audit:PigCVP}{}\SplitAuditName{PigCVP} & New sessions of known animals & Concern & Same as \hyperlink{split-audit:PigAirwayPressure}{\SplitAuditName{PigAirwayPressure}}. \\
\hypertarget{split-audit:Plane}{}\SplitAuditName{Plane} & New physical models or capture sessions & Unclear & The photographs show die-cast plane models, 30 shape samples per class; the source does not say how many physical models were photographed. The description does not establish how physical models or photography sessions were assigned to the two sets. The test could therefore contain another view of a familiar model; that possibility is not verified from the metadata. \\
\hypertarget{split-audit:PowerCons}{}\SplitAuditName{PowerCons} & New days or later periods & Concern & Nearby days can share weather and routines, and no source says whether a later period was kept for testing. Separately, 67 of 180 test cases exactly match training cases (Table~\ref{tab:duplicates}). These matches are a concrete issue beyond the possible relationship between neighbouring days. \\
\hypertarget{split-audit:ProximalPhalanxOutlineAgeGroup}{}\SplitAuditName{ProximalPhalanxOutlineAgeGroup} & New patients & Unclear & Same as \hyperlink{split-audit:DistalPhalanxOutlineAgeGroup}{\SplitAuditName{DistalPhalanxOutlineAgeGroup}}. \\
\hypertarget{split-audit:ProximalPhalanxOutlineCorrect}{}\SplitAuditName{ProximalPhalanxOutlineCorrect} & New patients & Unclear & Same as \hyperlink{split-audit:DistalPhalanxOutlineAgeGroup}{\SplitAuditName{DistalPhalanxOutlineAgeGroup}}. \\
\hypertarget{split-audit:ProximalPhalanxTW}{}\SplitAuditName{ProximalPhalanxTW} & New patients & Unclear & Same as \hyperlink{split-audit:DistalPhalanxOutlineAgeGroup}{\SplitAuditName{DistalPhalanxOutlineAgeGroup}}. \\
\hypertarget{split-audit:RefrigerationDevices}{}\SplitAuditName{RefrigerationDevices} & New appliances or households & Unclear & Same as \hyperlink{split-audit:Computers}{\SplitAuditName{Computers}}. \\
\hypertarget{split-audit:Rock}{}\SplitAuditName{Rock} & New physical samples & Unclear & The original check found strongly similar pairs of sequences. That alone does not show that the same physical rock was measured twice. We need sample identifiers to know whether the two sets share rocks. Until then, repeated measurements are a possibility, rather than confirmed overlap. \\
\hypertarget{split-audit:ScreenType}{}\SplitAuditName{ScreenType} & New appliances or households & Unclear & Same as \hyperlink{split-audit:Computers}{\SplitAuditName{Computers}}. \\
\hypertarget{split-audit:SemgHandGenderCh2}{}\SplitAuditName{SemgHandGenderCh2} & New people & Concern & The original audit reports the same five people in both sets, checked using labels from the related person-identification dataset. Recognising a familiar person could help predict their gender label. The test checks further measurements of these people, rather than establishing performance on people absent from training. \\
\hypertarget{split-audit:SemgHandMovementCh2}{}\SplitAuditName{SemgHandMovementCh2} & New people or recording occasions & Concern & The audit reports repeated grasps by the same people in the same recording sessions across both sets. These repetitions share the person and electrode placement. The test does not show whether recognition works for a new person or after removing and attaching the electrodes again. Those are separate evaluation settings. \\
\hypertarget{split-audit:SemgHandSubjectCh2}{}\SplitAuditName{SemgHandSubjectCh2} & New recording occasions & Concern & The task identifies people, but each person was recorded on one occasion according to the audit. Both sets contain repetitions from that occasion. The model might recognise that recording setup along with the person. The test does not establish identification after measuring the person again on another occasion. \\
\hypertarget{split-audit:ShakeGestureWiimoteZ}{}\SplitAuditName{ShakeGestureWiimoteZ} & New sessions of known users & Unclear & Same as \hyperlink{split-audit:PickupGestureWiimoteZ}{\SplitAuditName{PickupGestureWiimoteZ}}. \\
\hypertarget{split-audit:ShapeletSim}{}\SplitAuditName{ShapeletSim} & New generated examples & Generated & This is a generated task rather than measurements from people, devices or specimens. The corresponding source-separation question is therefore not assessed here. Its split would instead need to be examined through the generation procedure; synthetic origin alone is not evidence of leakage. \\
\hypertarget{split-audit:ShapesAll}{}\SplitAuditName{ShapesAll} & New source images or objects & Concern & The cases are outlines derived from shape images. The documentation does not establish whether repeated views or versions of one source object were kept in one set. Independence at the original image or object level remains unknown; no particular shared object is verified here. Separately, the value check reports two test cases identical to training data (Table~\ref{tab:duplicates}). \\
\hypertarget{split-audit:SmallKitchenAppliances}{}\SplitAuditName{SmallKitchenAppliances} & New appliances or households & Unclear & Same as \hyperlink{split-audit:Computers}{\SplitAuditName{Computers}}. \\
\hypertarget{split-audit:SmoothSubspace}{}\SplitAuditName{SmoothSubspace} & New generated examples & Generated & This is a generated task rather than measurements from people, devices or specimens. The corresponding source-separation question is therefore not assessed here. Its split would instead need to be examined through the generation procedure; synthetic origin alone is not evidence of leakage. \\
\hypertarget{split-audit:SonyAIBORobotSurface1}{}\SplitAuditName{SonyAIBORobotSurface1} & New recording occasions & Concern & The audit describes short pieces from one continuous robot recording per surface. Training and testing therefore share the robot run and its recording conditions. The test checks more pieces from those runs. It does not show whether surface recognition works after collecting another recording on a separate occasion. \\
\hypertarget{split-audit:SonyAIBORobotSurface2}{}\SplitAuditName{SonyAIBORobotSurface2} & New recording occasions & Concern & Same as \hyperlink{split-audit:SonyAIBORobotSurface1}{\SplitAuditName{SonyAIBORobotSurface1}}. \\
\hypertarget{split-audit:StarLightCurves}{}\SplitAuditName{StarLightCurves} & New stars & Separated & The source-level record describes one separate star per series. On that description, the test contains stars absent from training, rather than another piece of a familiar star's recording. No shared-star concern is identified. Possible shared observing conditions or processing procedures were not separately checked. \\
\hypertarget{split-audit:Strawberry}{}\SplitAuditName{Strawberry} & New physical samples & Concern & The test set is described as the source study's independently collected test data. However, the value check finds one sequence present in both sets with different labels (Table~\ref{tab:duplicates}). Thus, the original split description does not resolve this inconsistency in the released data. \\
\hypertarget{split-audit:SwedishLeaf}{}\SplitAuditName{SwedishLeaf} & New physical specimens & Concern & The cases are outlines derived from leaf images. The description does not establish how physical specimens and any repeated views were assigned to the two sets. New-specimen evaluation therefore cannot be verified; repeated views are a possibility to check, not an observed overlap. Separately, the value check reports two test cases identical to training data (Table~\ref{tab:duplicates}). \\
\hypertarget{split-audit:Symbols}{}\SplitAuditName{Symbols} & New people & Unclear & The source contains several symbol-drawing attempts per person. We do not know whether all attempts from a person were kept in one set. A test attempt could therefore come from a person already represented in training. This possible sharing needs participant identifiers to check; performance on new people is not established. \\
\hypertarget{split-audit:SyntheticControl}{}\SplitAuditName{SyntheticControl} & New generated examples & Generated & This is a generated task rather than measurements from people, devices or specimens. The corresponding source-separation question is therefore not assessed here. Its split would instead need to be examined through the generation procedure; synthetic origin alone is not evidence of leakage. \\
\hypertarget{split-audit:ToeSegmentation1}{}\SplitAuditName{ToeSegmentation1} & New people or recording occasions & Concern & The source contains repeated motion recordings from a small group of actors. The original audit suspects that actors are shared between the sets, but the files lack actor identifiers for checking this. We therefore cannot establish performance on new people or new recording occasions, and should not present actor overlap as verified. Separately, the value check finds four test cases identical to training data (Table~\ref{tab:duplicates}). \\
\hypertarget{split-audit:ToeSegmentation2}{}\SplitAuditName{ToeSegmentation2} & New people or recording sessions & Unclear & Same uncertainty about the actors and recording occasions as \hyperlink{split-audit:ToeSegmentation1}{\SplitAuditName{ToeSegmentation1}}. The four value matches mentioned there concern ToeSegmentation1, not this dataset. \\
\hypertarget{split-audit:Trace}{}\SplitAuditName{Trace} & New simulated events & Separated & Each series is described as a separate simulated event, so testing uses other generated events rather than another piece of one recording. No shared-event issue is identified by that description. Whether the generation process creates further dependencies would need its settings or code to check. \\
\hypertarget{split-audit:TwoLeadECG}{}\SplitAuditName{TwoLeadECG} & Independent recordings & Concern & The label identifies which ECG lead produced the signal. Training and test beats come from one patient's recording. The model may recognise differences specific to those leads during that measurement. The test does not show that the same distinction works when another patient or recording is used. \\
\hypertarget{split-audit:TwoPatterns}{}\SplitAuditName{TwoPatterns} & New generated examples & Generated & This is a generated task rather than measurements from people, devices or specimens. The corresponding source-separation question is therefore not assessed here. Its split would instead need to be examined through the generation procedure; synthetic origin alone is not evidence of leakage. \\
\hypertarget{split-audit:UMD}{}\SplitAuditName{UMD} & New generated examples & Generated & This is a generated task rather than measurements from people, devices or specimens. The corresponding source-separation question is therefore not assessed here. Its split would instead need to be examined through the generation procedure; synthetic origin alone is not evidence of leakage. \\
\hypertarget{split-audit:UWaveGestureLibraryAll}{}\SplitAuditName{UWaveGestureLibraryAll} & New days or recording occasions & Unclear & The original system recognises gestures for familiar users, so sharing users can be appropriate. The missing information is whether all recordings from a day were kept together. Without this, the test could contain gestures recorded on a day already represented in training, rather than gestures collected when the user returns. \\
\hypertarget{split-audit:UWaveGestureLibraryX}{}\SplitAuditName{UWaveGestureLibraryX} & New sessions or days & Unclear & Same as \hyperlink{split-audit:UWaveGestureLibraryAll}{\SplitAuditName{UWaveGestureLibraryAll}}. \\
\hypertarget{split-audit:UWaveGestureLibraryY}{}\SplitAuditName{UWaveGestureLibraryY} & New sessions or days & Unclear & Same as \hyperlink{split-audit:UWaveGestureLibraryAll}{\SplitAuditName{UWaveGestureLibraryAll}}. \\
\hypertarget{split-audit:UWaveGestureLibraryZ}{}\SplitAuditName{UWaveGestureLibraryZ} & New sessions or days & Unclear & Same as \hyperlink{split-audit:UWaveGestureLibraryAll}{\SplitAuditName{UWaveGestureLibraryAll}}. \\
\hypertarget{split-audit:Wafer}{}\SplitAuditName{Wafer} & New wafers or production runs & Concern & The archive ships the six sensor traces of each wafer as separate cases and splits them without regard to wafer or run. Matching every case to the raw files shows 690 wafers and 99 of 104 production runs on both sides; every training case has another sensor of its wafer in test. The test therefore does not measure performance on new wafers or runs. \\
\hypertarget{split-audit:Wine}{}\SplitAuditName{Wine} & New physical samples & Unclear & We do not know how many physical wines produced the spectral measurements or whether one wine was measured repeatedly. The two sets might contain measurements of the same wine. Not finding obvious similar pairs does not settle this question. Physical-sample identifiers are needed to check the split. \\
\hypertarget{split-audit:WordSynonyms}{}\SplitAuditName{WordSynonyms} & New manuscript pages & Unclear & Same as \hyperlink{split-audit:FiftyWords}{\SplitAuditName{FiftyWords}}. \\
\hypertarget{split-audit:Worms}{}\SplitAuditName{Worms} & New worms or recordings & Concern & The original record describes each case as a separate worm recording. However, the value check finds 10 of 77 test cases identical to training cases (Table~\ref{tab:duplicates}). We therefore need worm and recording identifiers to resolve the disagreement between that source description and the released values. \\
\hypertarget{split-audit:WormsTwoClass}{}\SplitAuditName{WormsTwoClass} & New worms or acquisitions & Concern & Same as \hyperlink{split-audit:Worms}{\SplitAuditName{Worms}}. \\
\hypertarget{split-audit:Yoga}{}\SplitAuditName{Yoga} & New filming occasions & Concern & The task identifies two actors from video frames. No test frame is identical to a training frame, but every training frame has a near-twin in test (median distance 2.09 versus 23.8 for random pairs), so both sets come from one pool of frames. The model may recognise familiar filming conditions along with the actor. The test does not establish identification in footage collected on another occasion. \\
\end{longtable}
\endgroup